# Perivascular space Identification Nnunet for Generalised Usage (PINGU)

Benjamin Sinclair[1*], Lucy Vivash[1,5,8], Jasmine Moses[1], Miranda Lynch[1], William Pham[1], Karina Dorfman[1], Cassandra Marotta[1], Shaun Koh[1], Jacob Bunyamin[1], Ella Rowsthorn[1], Alex Jarema[2], Himashi Peiris[3,7], Zhaolin Chen[3,7], Sandy R Shultz[1,5,6], David K Wright[1], Dexiao Kong[4,9,10], Sharon L. Naismith[4,9,10], Terence J. O'Brien[1,5,6], Meng Law[1,2]

[1] Department of Neuroscience, The School of Translational Medicine, Monash University, Melbourne, Australia

[2] Department of Radiology, Alfred Health, Melbourne, Australia

[3] Monash Biomedical Imaging. Monash University, Melbourne, Australia

[4] School of Psychology, Faculty of Science, University of Sydney, Sydney, Australia

[5] Department of Neurology, Alfred Health, Melbourne, Australia

[6] Centre for Trauma and Mental Health, Vancouver Island University, Nanaimo, Canada

[7] Department of Data Science and AI, Monash University, Melbourne, Australia

[8] Department of Medicine, The Royal Melbourne Hospital, The University of Melbourne, Parkville, Australia.

[9] Healthy Brain Ageing Program, Brain and Mind Centre, University of Sydney, Camperdown, NSW 2050, Australia

[10] Charles Perkins Centre, University of Sydney, Camperdown, NSW 2050, Australia

*Corresponding Author: Ben Sinclair: ben.sinclair@monash.edu

Address: Department of Neuroscience, Monash University, 99 Commercial Road, Melbourne, Australia, 3004

Abstract

Perivascular spaces (PVSs) form a central component of the brain's waste clearance system, the glymphatic system. These structures are visible on MRIs when enlarged, and their morphology is associated with aging and neurological disease. Manual quantification of PVS is time consuming and subjective. Numerous deep learning methods for PVS segmentation have been developed for automated segmentation. However, the majority of these algorithms have been developed and evaluated on homogenous datasets and high resolution scans, perhaps limiting their applicability for the wide range of image qualities acquired in clinical and research settings. In this work we train a nnUNet, a top-performing task driven biomedical image segmentation deep learning algorithm, on a heterogenous training sample of manually segmented MRIs of a range of different qualities and resolutions from 6 different datasets acquired on 5 different scanners. These are compared to the two currently publicly available deep learning methods for 3D segmentation of PVS, evaluated on scans with a range of resolutions and qualities. The resulting model, PINGU (Perivascular space Identification Nnunet for Generalised Usage), achieved voxel and cluster level dice scores of 0.50(SD=0.15) and 0.63(0.17) in the white matter (WM), and 0.54 (0.11) and 0.66(0.17) in the basal ganglia (BG). Performance on unseen data was substantially lower for both PINGU (0.20-0.38 [WM, voxel], 0.29-0.58 [WM, cluster], 0.22-0.36 [BG, voxel], 0.46-0.60 [BG, cluster]) and the publicly available algorithms (0.18-0.30 [WM, voxel], 0.29-0.38 [WM cluster], 0.10-0.20 [BG, voxel], 0.15-0.37 [BG, cluster]). Nonetheless, PINGU strongly outperformed the publicly available algorithms, particularly in the BG. Finally, training PINGU on manual segmentations from a single site with homogenous scan properties (as opposed to training on the larger, multi-site dataset) gave marginally lower performances on internal cross-validation, but in some cases gave higher performance on external validation on unseen sites data. PINGU stands out as broad-use PVS segmentation tool, with particular strength in the BG, an area of PVS highly related to vascular disease and pathology.

1. Introduction

Perivascular spaces (PVSs) are emerging as an important biomarker in neurological disease and ageing. They form part of the glymphatic system, and assist in removal of waste from the extracellular space of the brain parenchyma (Mestre et al., 2020; Wardlaw et al., 2020). Reduced waste clearance is suggested to result in the enlargement of PVS that become visible on T1- and T2-weighted MRI (Xue et al., 2020). Traditionally, enlarged PVSs were identified and counted manually on representative MRI slices, and converted to a scale representing burden (Potter et al., 2015). Now, a number of automated methods are available to mark voxels containing PVS on 3D MRI images. These include filtering-based methods (Sepehrband et al., 2019), intensity-based methods (Schwartz et al., 2019), and machine learning based methods (Boutinaud et al., 2021; Lian et al., 2018).

Machine learning algorithms can achieve impressive performance within a particular domain. Whilst algorithms are tested on unseen data, the test set typically occupies the same domain as the training data, thus they often fail to perform as well on data from a different domain (i.e. they do not generalise well). In the context of MRI biomedical segmentation, examples of different domains are different scanners, field strengths, age groups, disease groups, and voxel sizes which may impact the quality and comparability of results. One limitation of existing methods is that they are trained and validated on homogenous data sets from a single site with a single MRI protocol. To improve generalisability to new datasets, the algorithms should be trained and validated on a variety of data representing the variability likely to be encountered in the algorithm's application. In particular, PVS characteristics on MRI differ considerably with age (Bown et al., 2022; Lynch et al., 2023), and disease status (Lynch et al., 2022; Wardlaw et al., 2020), so it is important that deep learning algorithms have been trained on a range of age and disease groups.

The UNet is a deep learning algorithm developed in 2015 for the segmentation of biomedical images (Ronneberger et al., 2015). SHIVA-PVS (Boutinaud et al., 2021) and M$^2$EDN (multi-channel multi-scale deep convolutional encoder-decoder network) (Lian et al., 2018) used a UNet-based architecture in their algorithms to segment PVSs. More recently, the nnUNet (Isensee et al., 2021) was developed in 2020 and is self-configuring, meaning that multiple different UNet architectures and hyperparameters are automatically selected to give optimal performance. The nnUNet is now the mainstay in biomedical image segmentation, however it has not yet been applied to the segmentation of PVSs on 3D images.

This work makes two novel contributions to the field. Firstly, we trained and validated the algorithm with a diverse dataset from 7 different studies from 6 different scanners, which included an evaluation of performance on data from unseen sites. Secondly, we optimised and applied the nnUNet to the problem of 3D PVS segmentation, which has previously been done for 2D images (Sudre et al., 2024), but not for 3D images, and compared its performance on data from unseen sites to currently publicly-available deep learning methods for PVS segmentation (SHIVA-PVS (Boutinaud et al., 2021) and WPSS (Lan et al., 2023))

2. Methods

*2.1 Participants*

Data came from participants of 4 studies with neuroimaging from Australia, and 3 publicly available datasets from the U.S.A. The Australian studies were from the Healthy Brain Ageing Program (HBA) at the Brain and Mind Centre, University of Sydney (Kong et al., 2021), and contains older adults 'at risk' for dementia (namely those with Mild Cognitive

Impairment or with early Alzheimer's Disease, Vascular Disease or mixed dementias), the Australian Football study (AF), Royal Melbourne Hospital of amateur Australian rules football (Major et al., 2021) , the Fronto-Temporal Dementia study (FTD), Royal Melbourne Hospital, of patients enrolled in an open label phase 1b study of sodium selenate as a treatment for fronto-temporal dementia (Vivash et al., 2022) and the mild cognitive impairment study (MCIS), Alfred Hospital, characterising cognitive and biomarker profiles of patients with suspected mild cognitive impairment (ACTRN12620001246976). The publicly available datasets were the Alzheimer's Disease Neuroimaging Initiative (ADNI) (Jack Jr et al., 2008), The Human Connectome Project (HCP) (Van Essen et al., 2013), and the Nathan-Kline Rockland Sample (NKI-RS) (Nooner et al., 2012). The HCP and NKI-RS datasets were combined into an assorted controls (ASC) dataset as only 2 images were used for each of these datasets, and they represent the same demographic. Further details of each study can be found in the corresponding citations. The only inclusion criteria was the availability of 3D T1-weighted (T1w) MRI imaging.

From each site, a variable number of participants were randomly selected. Table 1 gives details of the trials, subject demographics, and MRI parameters from each trial.

*2.2 Manual Segmentation of PVS*

Manual PVS labels were marked using ITKSNAP (version 3.3.0). Seven raters (JM, ML1, WP, KD, CM, SK, JB) were trained by experienced neuroradiologist (ML2) and imaging scientist (BS) on how to identify PVS and distinguish them from similarly appearing imaging features such as lacunes, white matter hyperintensities, and sulci. For each cohort, a single rater segmented all voxels they deemed to be within PVSs. A second rater then checked the first rater's segmentations and added in any voxels missed, and queried any voxels disagreed upon, which would be settled by a third senior neuroradiologist rater (ML2, AJ). The two largest datasets (ADNI and AF) were all checked by a senior third rater (ML2, AJ).

Table 1: Description of the 6 datasets used for manual segmentation and algorithm training and evaluation. ADNI: Alzheimer's Disease Neuroimaging Initiative, AF: Australian Football study, ASC: Assorted Controls, FTD: Frontotemporal Dementia, MCIS: Mild Cognitive Impairment, AD: Alzheimer's Disease, CN: Cognitively Normal, WM: White Matter, BG: Basal Ganglia. RMH: Royal Melbourne Hospital. Numeric values presented as Mean (SD).

| Study | ADNI | AF | ASC | HBA | FTD | MCIS |
|---|---|---|---|---|---|---|
| n | 10 | 10 | 4 | 5 | 4 | 7 |
| Location | USA - various | RMH | USA - various | IMED Radiology (Camperdown) | RMH | Alfred |
| Age | 74.1 (4.1) | 26.2 (1.8) | 21 (8.4) | 64.4 (8.3) | 61.3 (10.1) | 68.6(6.4) |
| Condition | CN (3) MCI (4) AD (3) | AF players (4) AF concussion (1) Non-contact sport (5) | Healthy Controls | Subjective Cognitive Decline (1) MCI (4) | Behavioural variant FTD | FTD (1) Parkinson's-plus syndrome (1) MCI-AD spectrum(5) |
| Scanner | Signa Excite(5) Symphony(3) Genesis Signa(1) Trio(1) | Siemens Magnetom Prisma | Siemens (Custom) Siemens TrioTim | General Electric Discovery MR750 | Siemens Magnetom Prisma | Siemens Magnetom Prisma |
| Field Strength | 1.5T | 3T | 3T | 3T | 3T | 3T |
| Sequence | MPRAGE | MPRAGE | MPRAGE | BRAVO SPGR | MPRAGE | MPRAGE |
| Voxel size | 0.9x0.9x1.2(5) 1.0x1.0x1.2(1) 1.1x1.1x1.2(1) 1.3x1.3x1.2(3) | 0.8x0.8x0.8 | 1.0x1.0x1.0(2) 1.0x0.98x0.98(2) | 1.0x1.0x1.0 | 1.0x1.0x1.0 | 0.8x0.8x0.8 |
| Average number PVS (WM) | 166 (92) | 266 (135) | 218 (67) | 303 (169) | 292 (124) | 484 (311) |
| Average volume PVS (WM) | 3580 (1491) | 3162 (2023) | 2962 (1161) | 1943 (1394) | 1703 (538) | 2130 (1453) |
| Average number PVS (BG) | 32 (9) | 33 (9) | 17 (8) | 66 (27) | 50 (10) | 98 (41) |
| Average volume PVS (BG) | 1347 (617) | 957 (235) | 294 (154) | 386 (327) | 476 (137) | 385 (176) |

*2.3 Pre-processing*

nnUNet's default pre-processing was carried out with only one modification. Rather than allow the nnUNet to select the "target spacing" (resampled resolution) of input images itself, as a median over the datasets, we set the target spacing to 0.8mm$^3$, that is the highest resolution amongst our different datasets. This is because PVSs are fine structures, often at sub-millimetre resolution, and any down-sampling of any of the datasets would reduce their observability. T1w images were intensity-normalised to a z-score, resampled to the specified target spacing, and in-training data-augmentation included: rotations, scaling, Gaussian noise, Gaussian blur, brightness, contrast, simulation of low resolution, gamma correction and mirroring.

*2.3 Deep Learning*

The nnUNet (Isensee et al., 2021) version 1.7.0 was used to train a model to identify PVS. The nnUNet is based on a UNet convolutional neural network architecture (Ronneberger et al., 2015), which consists of encoder and decoder pathways, where the encoder successively reduces the dimensionality of representations of the input image via convolution operations, thereby forcing the network to learn an efficient representation containing the most salient features, and the decoder expands these representations back to a full resolution image via deconvolutions. The UNet's innovation was in its skip connections between equivalent levels of the encoder and decoder, which transfers information from the different spatial scales to the decoder, allowing the network to learn both small-scale and large-scale features of the object classes. The nnUNet is a self-configuring UNet framework, whereby the model architecture (depth, number of convolution channels etc.), hyperparameters, and data augmentation, are all either selected automatically, or optimised based on the dataset. It has field leading performance across a range of different image segmentation tasks (Isensee et al., 2021).

*2.4.1 Internal Evaluation*

We utilised a 5-fold cross-validation (5FCV) for our first training and evaluation schedule. This training schedule is the most commonly reported method in deep learning studies of

PVS segmentation and is expected to result in a model with high performance and generalisability on unseen sites since the model has seen examples of a wide range of datasets and will be less prone to overfitting to the characteristics of a particular dataset. The data is split into 5 folds each containing 20% of the data. The model is then trained on 4 of the folds and assessed on the 5$^{th}$ fold, which is repeated with each fold being left out from training as the validation fold once. By default, the nnUNet both trains and evaluates using this 5FCV procedure, such that there are 5 trained models output, each trained on a different 80% of the data. These 5 models can then be combined to run inference on new data.

*2.4.2 External Evaluation*

To assess how the model performs on images from an unseen dataset with different properties to the training data, the second training + evaluation schedule was a leave-one-site-out cross-validation (LOSOCV), whereby the nnUNet was trained on all data from 5 sites, and its performance evaluated on the 6$^{th}$ site. This was then repeated leaving out each site successively, such that each subject would be in the evaluation set for one and only one of these runs. The performance measures where then averaged across subjects. This training schedule will provide a good estimation of the algorithms' out-of-sample performance when applied to new data.

*2.4.3 Single Site Training*

The final training + evaluation schedule was where the model was trained on data from a single site and evaluated on the remaining 5 sites. This was undertaken for two reasons. Firstly to see if training on multiple datasets, as per the LOSOCV, does indeed improve generalisability and performance in unseen sites' data. Secondly, for comparison with the other two publicly available algorithms, and the literature more broadly, which generally trains and evaluates on data from a single site, with consistent scanning parameters. This procedure was only conducted for sites with at least 5 subjects per site (ADNI, AF, CGS, MCIS), since the nnUNet needs at least 5 images to perform its 5FCV training procedure.

*2.5 Post-processing*

To assess PVS volumes and numbers in regions of interest, each subject's T1w image was non-linearly registered to the MNI template space provided by FSL (Jenkinson et al., 2012) using ANTS Syn registration (Avants et al., 2008). The inverse deformation fields were used to propagate atlas segmentations of WM and the Basal Ganglia. The WM was extracted from the Harvard-Oxford Cortical atlas (Desikan et al., 2006). The Basal Ganglia region included the Putamen, Pallidum, Caudate, Nucleus Accumbens, and Thalamus extracted from the Harvard-Oxford Sub-cortical atlas along with the internal capsule (posterior and anterior limbs) extracted from the JHU DTI-based WM atlas (Mori et al., 2005). Subsequent model performance measures were calculated separately for each region.

*2.6 Model Assessments*

For each of the model variations described in the previous section, our primary performance measure was the voxel-level dice score ($DSC_{vox}$), a measure of overlap between ground truth and algorithm segmentation, which was calculated for each subject in the validation set(s) and averaged over subjects. Additionally, we quantified voxel level sensitivity ($Sen_{vox}$), positive predictive value ($PPV_{vox}$), and correlation ($r_{vox}$) across subjects between the volume of manually segmented PVS voxels and volume of algorithm segmented voxels. Correlations were only calculated in the three largest datasets with n≥7. Further, correlations were averaged across datasets rather than across subjects, since the behaviour of the algorithms were different in each dataset, so we did not want to combine subjects from different datasets into a single correlation. Since it is common in the literature to use counts of MRI-visible PVS as a marker of enlarged PVS burden, the count-level (also referred to as cluster-level) equivalents of each of these performance measures was also calculated. For this, unique PVS clusters were identified as all distinct connected segmented voxels, via the skimage.measure.labels clustering algorithm in python's skimage package. A cluster was considered a true positive if any voxel within the algorithm-segmented cluster overlapped with any voxel from the manual segmentation, and a false positive if it did not. Definitions of each of these outcome measures are given in Table 2.

Table 2: Definition of performance measures. n=number of distinct clusters in segmentation

| Performance Measure | Voxel Level | Cluster Level |
|---|---|---|
| Dice Score | $DSC_{vox} = \dfrac{2 \; x \; volume(overlap)}{volume(manual) + volume(algo)}$ | $DSC_{num} = \dfrac{2 \; x \; n(overlapping)}{n(manual) + n \; (algo)}$ |
| Sensitivity (Recall) | $Sen_{vox} = \dfrac{volume(overlap)}{volume(manual)}$ | $Sen_{num} = \dfrac{n(overlap)}{n(manual)}$ |
| Positive Predictive Value (Precision) | $PPV_{vox} = \dfrac{volume(overlap)}{volume(algo)}$ | $PPV_{num} = \dfrac{n(overlap)}{n(algo)}$ |
| Correlation | $r_{vox} = r(volume(manual), volume(algo))$ | $r_{num} = r(n(manual), n(algo))$ |

*2.7 Comparator Algorithms*

We compared the nnUNet models to existing models in the literature that were publicly available at the time of writing. A recent systematic review (Waymont et al., 2024) identified nineteen deep learning algorithms for identifying PVS. Of these, eight (Abbruzzese et al., 2016; Boutinaud et al., 2021; Choi et al., 2020; Lian et al., 2018; Rashid et al., 2023; Sudre et al., 2024) output a segmentation map, whilst the other eleven output a PVS rating scale/score, enhanced image, distance map or a lesion heatmap rather than a segmented image (Table 3). Of the eight segmentation algorithms, four (Choi & Jin, 2018; Choi et al., 2020; Lan et al.,

2023; Lian et al., 2018) worked on 3D images, as opposed to 2D slices (Rashid et al., 2023; Sudre et al., 2024), and only two of the four were publicly available. The first was another UNet based method, SHIVA-PVS (Boutinaud et al., 2021), and the second, Weakly Supervised Perivascular Space Segmentation (WPSS) (Lan et al., 2023) used a Frangi filter (Frangi et al., 1998) to highlight vessel like structures, then a UNet to remove false positives. Both of these algorithms were used to segment all 40 of the T1w images in our dataset, and the performance measures calculated on each of the datasets.

Table 3: Availability of deep learning PVS segmentation algorithms (adapted from (Waymont et al., 2024).

| Study | Outcome Features | Dimension | Publicly Available |
|---|---|---|---|
| (Dubost et al., 2017) | Lesion heatmaps | - | - |
| (Jung et al., 2018) | Enhanced T2w image | - | - |
| (Dubost, Adams, et al., 2019) | PVS scores and saliency maps (in BG only) | - | - |
| (Dubost, Yilmaz, et al., 2019) | PVS scores in 4 regions | - | - |
| (Dubost, Dünnwald, et al., 2019) | PVS scores and attention maps | - | - |
| (Jung et al., 2019) | Enhanced image | - | - |
| (Sudre et al., 2019) | Probabilistic boxes identifying presence of PVS and lacunes | - | - |
| (van Wijnen et al., 2019) | Distance map | - | - |
| (Dubost et al., 2020) | PVS detection, segmentation, volume (and num) in 4 regions | - | - |
| (Yang et al., 2021) | BG PVS scores | - | - |
| (Williamson et al., 2022) | PVS scores | - | - |
| (Rashid et al., 2023) | PVS segmentation | 2D | - |
| (Sudre et al., 2024) 'BigrBrain' | PVS detection and segmentation, Total PVS volume & num | 2D | - |
| (Sudre et al., 2024) 'TeamTea' | PVS detection and segmentation, Total PVS volume & num | 2D | - |
| (Sudre et al., 2024) 'Neurophet' | PVS detection and segmentation, Total PVS volume & num | 2.5D | - |
| (Lian et al., 2018) | PVS segmentation | 3D | No |
| (Choi et al., 2020) | PVS segmentation and regional (BG and WM) volumes | 3D | No |
| (Boutinaud et al., 2021) | PVS segmentation and regional volumes in BG and deep WM | 3D | Yes |
| (Lan et al., 2023) | PVS segmentation, saliency map | 3D | Yes |

2.7.1 SHIVA

SHIVA-PVS version 1 (enhanced from publication) was used, available from https://github.com/pboutinaud/SHIVA_PVS. SHIVA-PVS requires some pre-processing prior to application of the algorithm. Images must be resampled to 1mm resolution, cropped to (160x214x176) voxels, and normalised between 0-1. T1w images were reoriented to MNI standard space using fsl reorient to standard and resampled to 1mm isotropic resolution. The extent of the brain was determined using AFNIs 3dSkullStrip, and the x,y and z coordinates in the middle of that range extracted. A volume of (160x214x176) was extracted centred on these mid-range x,y and z coordinates. The 99th percentile intensity was extracted, and voxels scaled to 0-1 within this range, with all voxels above the 99th percentile set to 1, as per (Boutinaud et al., 2021). SHIVA-PVS was then applied to this image using default parameters. Since the resulting segmentation is in the extracted volume space, this volume was resampled to the participants native T1w space using spm12's (Penny et al., 2011) resample function, with nearest neighbour interpolation as appropriate for binary images.

*2.7.2 WPSS*

WPSS segments images in the native space of the input image, and no further processing was required. The code is available at https://github.com/Haoyulance/WPSS, and model weights are available from the authors on request for "Enhanced Perivascular Space Contrast" (a combination of T1w and T2w images), and T1w only, which was used for this study.

*2.8 Data and Model Availability*

Our code and model weights are publicly available at: https://github.com/iBrain-Lab/PINGU. We share 3 sets of model weights, 1) PINGU-All trained on all sites, 2) PINGU-AF trained on AF only, which performed best in the WM on unseen sites, and 3) PINGU-ADNI trained on ADNI only, which performed best in the BG on unseen sites. The code used to pre-process the data for SHIVA-PVS is available on the same web-page. Due to patient data privacy regulations, only the manual segmentations from publicly available ADNI, HCP and NKI-RS are made available (corresponding T1w images available from the respective studies' websites). The other data used to train this model can not be made publicly available, but may be provided upon request to qualified researchers subject to approval by the local ethics committee.

## 3. Results

### 3.1 Internal Validation

Average dice scores, correlations, sensitivity and precision for voxels and numbers of PVS for the internal 5FCV are given in Table 4. More detailed results on how each algorithm performed in each dataset are given in supplementary tables S1-S8 and Figures 2-3. The average dice score 5-fold cross-validation using all sites' data was $DSC_{vox}$=0.50 (SD=0.15) in the WM and $DSC_{vox}$=0.54 (0.10) in the BG (Table 4, Figures 3a, 4a). The equivalent cluster-level dice score was $DSC_{num}$=0.63 (0.17) in the WM and $DSC_{num}$=0.66 (0.17) in the BG (Table 4, Figures 3b,4b). The sensitivities were $Sen_{vox}(WM)$=0.41 (0.16), $Sen_{vox}(BG)$=0.46 (0.15), $Sen_{num}(WM)$=0.55 (0.20), $Sen_{num}(BG)$=0.56 (0.17) whilst precision was $PPV_{vox}(WM)$=0.75 (0.14), $PPV_{vox}(BG)$=0.74 (0.11), $PPV_{num}(WM)$=0.77 (0.14), $PPV_{num}(BG)$=0.76 (0.13). The average correlation across datasets between manual segmentation and PINGU segmentation was $r_{vox}(WM)$=0.80 (0.15), $r_{vox}(BG)$= 0.77 (0.11) for voxels and $r_{num}(WM)$=0.84 (0.14), $r_{num}(BG)$= 0.69 (0.15) for clusters (Table 4).

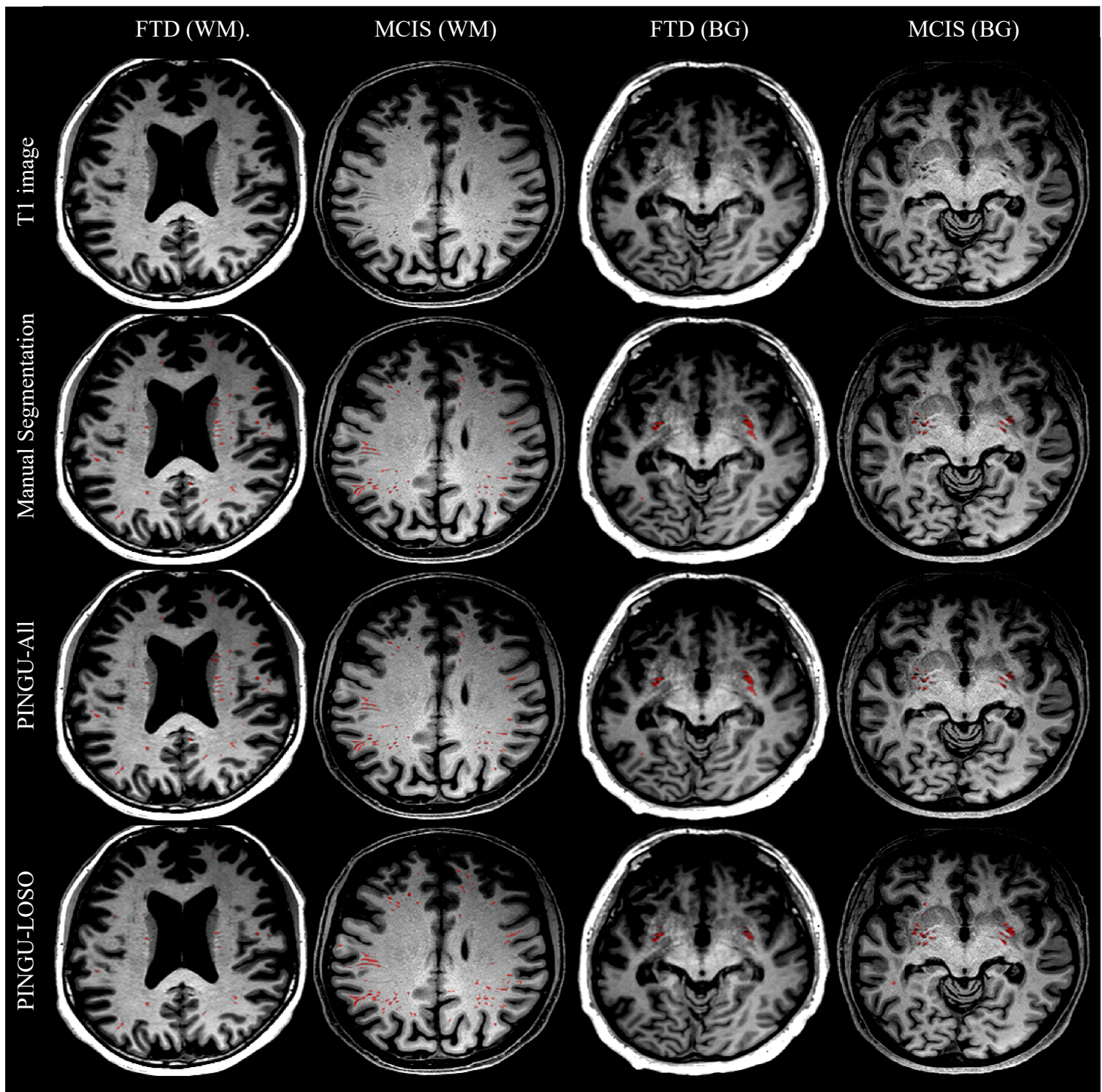

Figure 1: Overlay of PINGU-All and PINGU-LOSO on T1 image, and comparison to the manual segmentations, for one "high quality" dataset (MCIS), and one "low quality" dataset (FTD). Display slices were chosen automatically as those with the largest volume of PVS in WM and BG from the manual segmentations.

Table 4: Summary of average performance measures from internal validation for each training + evaluation schedule. Mean (standard deviation) presented, where mean is average over subjects for Dice Score, Sensitivity and Precision, and over datasets for correlation.

| Region | Algorithm | Training Dataset | Evaluation Dataset | Dice Score voxel | Dice Score number | Sensitivity voxel | Sensitivity number | Precision voxel | Precision number | Correlation voxel | Correlation number |
|---|---|---|---|---|---|---|---|---|---|---|---|
| White Matter | PINGU-All | All Sites | ADNI | 0.47 (0.13) | 0.70 (0.11) | 0.34 (0.12) | 0.60 (0.14) | 0.81 (0.10) | 0.80 (0.08) | 0.69 | 0.80 |
| | | | AF | 0.61 (0.09) | 0.68 (0.08) | 0.54 (0.13) | 0.65 (0.12) | 0.74 (0.14) | 0.71 (0.16) | 0.74 | 0.76 |
| | | | ASC | 0.56 (0.13) | 0.80 (0.09) | 0.42 (0.13) | 0.66 (0.13) | 0.89 (0.03) | 0.87 (0.06) | - | - |
| | | | CGS | 0.38 (0.19) | 0.34 (0.14) | 0.28 (0.15) | 0.23 (0.10) | 0.73 (0.14) | 0.72 (0.21) | - | - |
| | | | FTD | 0.34 (0.13) | 0.45 (0.16) | 0.23 (0.11) | 0.31 (0.14) | 0.77 (0.09) | 0.87 (0.09) | - | - |
| | | | MCIS | 0.55 (0.07) | 0.67 (0.09) | 0.52 (0.09) | 0.65 (0.09) | 0.60 (0.10) | 0.70 (0.15) | 0.97 | 0.96 |
| | | | All sites | 0.50 (0.15) | 0.63 (0.17) | 0.41 (0.16) | 0.55 (0.20) | 0.75 (0.14) | 0.77 (0.14) | 0.80 (0.15) | 0.84 (0.11) |
| | PINGU-ADNI | ADNI | ADNI | 0.45 (0.11) | 0.70 (0.09) | 0.33 (0.10) | 0.60 (0.12) | 0.80 (0.07) | 0.79 (0.10) | 0.74 | 0.80 |
| | PINGU-AF | AF | AF | 0.60 (0.08) | 0.65 (0.09) | 0.54 (0.13) | 0.65 (0.12) | 0.73 (0.15) | 0.68 (0.20) | 0.78 | 0.61 |
| | PINGU-CGS | CGS | CGS | 0.23 (0.13) | 0.25 (0.11) | 0.14 (0.08) | 0.15 (0.07) | 0.81 (0.14) | 0.80 (0.21) | - | - |
| | PINGU-MCIS | MCIS | MCIS | 0.51 (0.06) | 0.66 (0.09) | 0.42 (0.08) | 0.58 (0.08) | 0.68 (0.08) | 0.75 (0.14) | 0.98 | 0.99 |
| Basal Ganglia | PINGU-All | All Sites | ADNI | 0.51 (0.11) | 0.79 (0.11) | 0.40 (0.13) | 0.66 (0.08) | 0.77 (0.10) | 0.79 (0.10) | 0.73 | 0.74 |
| | | | AF | 0.66 (0.03) | 0.69 (0.08) | 0.60 (0.06) | 0.61 (0.09) | 0.74 (0.07) | 0.68 (0.13) | 0.65 | 0.52 |
| | | | ASC | 0.47 (0.10) | 0.75 (0.20) | 0.36 (0.11) | 0.63 (0.13) | 0.72 (0.12) | 0.70 (0.19) | - | - |
| | | | CGS | 0.41 (0.11) | 0.42 (0.15) | 0.29 (0.11) | 0.29 (0.13) | 0.80 (0.14) | 0.83 (0.11) | - | - |
| | | | FTD | 0.50 (0.07) | 0.49 (0.15) | 0.36 (0.08) | 0.34 (0.14) | 0.83 (0.04) | 0.85 (0.11) | - | - |
| | | | MCIS | 0.58 (0.06) | 0.64 (0.06) | 0.57 (0.14) | 0.62 (0.12) | 0.63 (0.09) | 0.74 (0.14) | 0.92 | 0.81 |
| | | | All sites | 0.54 (0.11) | 0.66 (0.17) | 0.46 (0.15) | 0.56 (0.17) | 0.74 (0.11) | 0.76 (0.13) | 0.77 (0.14) | 0.69 (0.15) |
| | PINGU-ADNI | ADNI | ADNI | 0.50 (0.08) | 0.77 (0.10) | 0.39 (0.10) | 0.66 (0.06) | 0.76 (0.08) | 0.74 (0.09) | 0.87 | 0.80 |
| | PINGU-AF | AF | AF | 0.66 (0.03) | 0.67 (0.09) | 0.61 (0.05) | 0.63 (0.10) | 0.73 (0.08) | 0.62 (0.11) | 0.53 | 0.22 |
| | PINGU-CGS | CGS | CGS | 0.28 (0.10) | 0.33 (0.11) | 0.18 (0.08) | 0.22 (0.12) | 0.80 (0.14) | 0.84 (0.15) | - | - |
| | PINGU-MCIS | MCIS | MCIS | 0.56 (0.07) | 0.63 (0.08) | 0.49 (0.12) | 0.57 (0.12) | 0.68 (0.10) | 0.76 (0.14) | 0.85 | 0.89 |

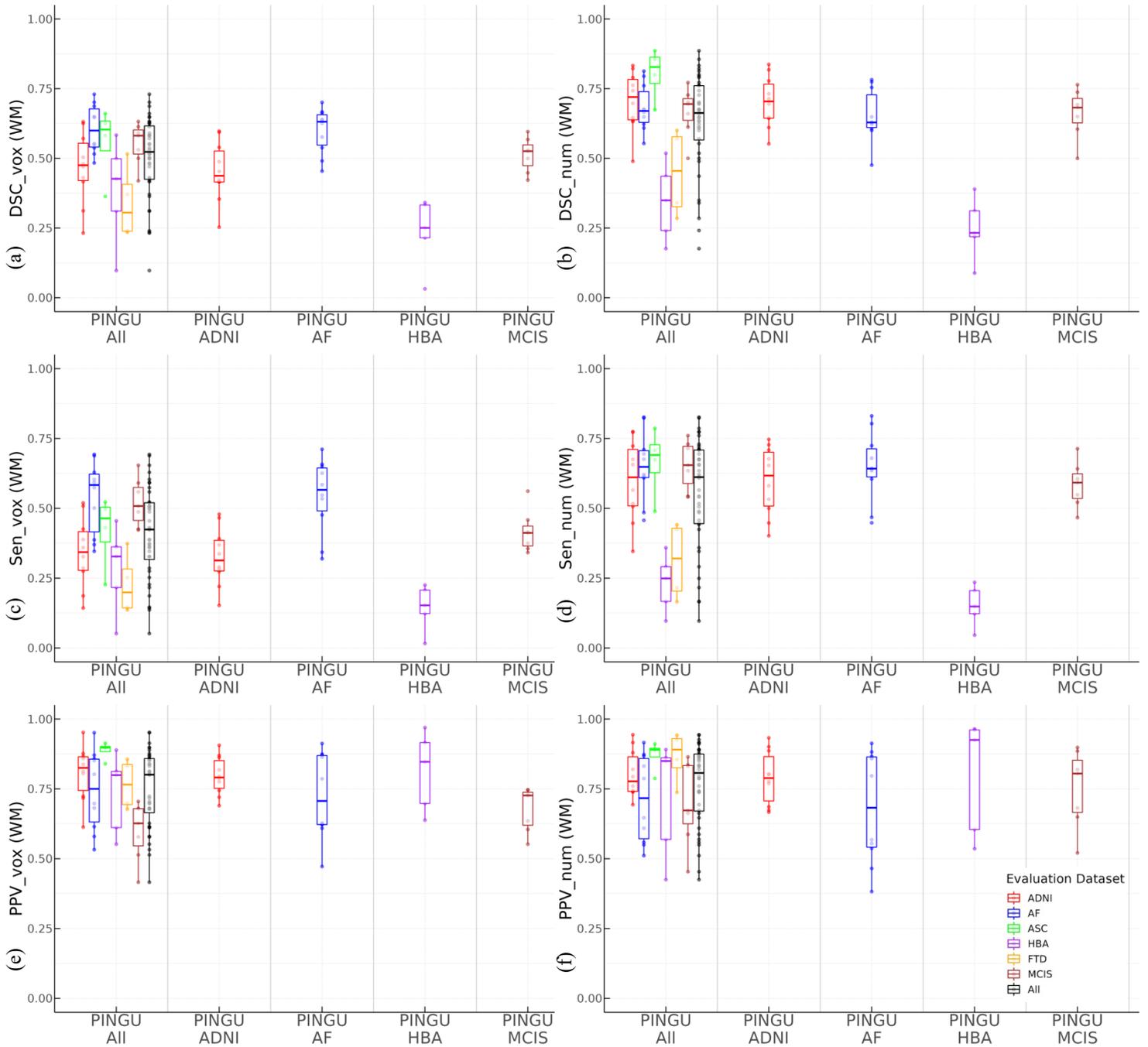

Figure 2: Internal validation (white matter (WM)): Performance for PINGU trained and evaluated on all sites' data, and on each dataset individually, in the white matter (WM). DSC: dice score, Sen: sensitivity, PPV: Positive Predictive Value/Precision, vox: voxel-level metrics, num: cluster-level metrics.

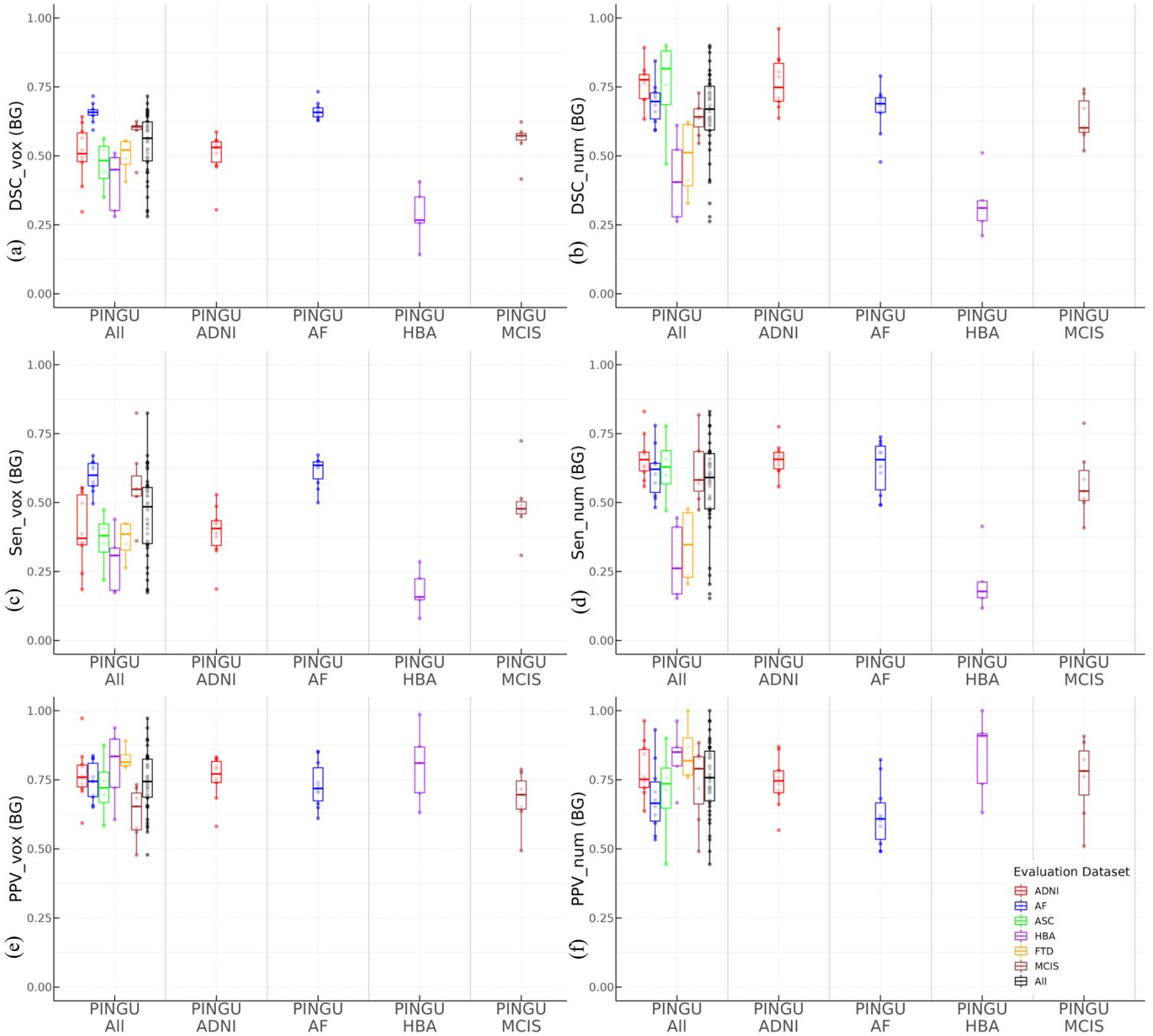

Figure 3: Internal validation (Basal Ganglia (BG)): Performance for PINGU trained and evaluated on all sites' data, and on each dataset individually. DSC: dice score, Sen: sensitivity, PPV: Positive Predictive Value/Precision, vox: voxel-level metrics, num: cluster-level metrics.

*3.2 External Validation and Algorithm Comparison*

In the Leave-One-Site-Out external validation, dice scores of PVS voxels were substantially lower (-24%(WM), -33%(BG)) in the unseen sites' data ($DSC_{vox}$(WM)=0.38(0.13), $DSC_{vox}$(BG)=0.36(0.13)) (Table 5) compared to the internal 5FCV ($DSC_{vox}$(WM)=0.50(0.15), $DSC_{vox}$(BG)=0.54(0.10)), though interestingly were only (-8%(WM), -9%(BG)) lower for PVS clusters ($DSC_{num}$(WM)=0.58(0.17), $DSC_{num}$(BG)=0.60(0.20)) (Table 5) compared to $DSC_{num}$(WM)=0.63(0.17) and $DSC_{num}$(BG)=0.66(0.17) for the internal 5FCV, indicating that in external sites' data, PINGU still detects a large proportion of the PVS, but does not fully segment each voxel of the PVSs.

In the unseen sites, PINGU strongly outperformed SHIVA ($DSC_{vox}$(WM)=0.18, $DSC_{vox}$(BG)=0.10, $DSC_{num}$(WM)=0.29, $DSC_{num}$(BG)=0.15) (Tables 5, S1) and WPSS ($DSC_{vox}$(WM)=0.30, $DSC_{vox}$(BG)=0.20, $DSC_{num}$(WM)=0.38, $DSC_{num}$(BG)=0.37) (Tables 5,S2), particularly in the BG. In terms of correlation with manual segmentations, all methods performed well in the WM: SHIVA: $r_{vox}$(WM)=0.80, WPSS: $r_{vox}$(WM)=0.86, PINGU: $r_{vox}$(WM)=0.82 (Table 5), whilst in the BG, only PINGU achieved a strong correlation: SHIVA: $r_{vox}$(BG)=0.11, WPSS: $r_{vox}$(WM)=0.56, PINGU: $r_{vox}$(WM)=0.78 (Table 5).

*3.3 Single Site Training*

Within the ADNI, AF, CGS and MCIS datasets, models trained with data from only ADNI (PINGU-ADNI), AF (PINGU-AF), CGS (PINGU-CGS) , and MCIS (PINGU-MCIS) yielded $DSC_{vox}$(WM) that were -4.3%,-1.6%,-39.5%,-7.3% lower, and DSCvox(BG) that were -2.0%,0.0%,-31.7%,-3.4% lower in their respective datasets compared to PINGU trained on all data (PINGU-All) (Table 5, Figures 2a, 3a, S1,S2). The highest internal 5FCV were from PINGU-AF (WM:0.60, BG:0.66).

In the external validation on unseen sites, PINGU-AF ($DSC_{vox}$(WM)=0.47, $DSC_{num}$(WM)=0.57) strongly outperformed PINGU-All ($DSC_{vox}$(WM)=0.38, $DSC_{num}$(WM)=0.58) in the WM on voxel-level dice score, but not cluster-level dice score (Table 5, Figures 7a,b,S1,S2), and PINGU-ADNI ($DSC_{vox}$(BG)=0.46, $DSC_{num}$(BG)=0.55) strongly outperformed PINGU-All ($DSC_{vox}$(BG)=0.36, $DSC_{num}$(BG)=0.60) in the BG on voxel level dice score, but not number level dice score (Table 5, Figures 8a,b,S1,S2).

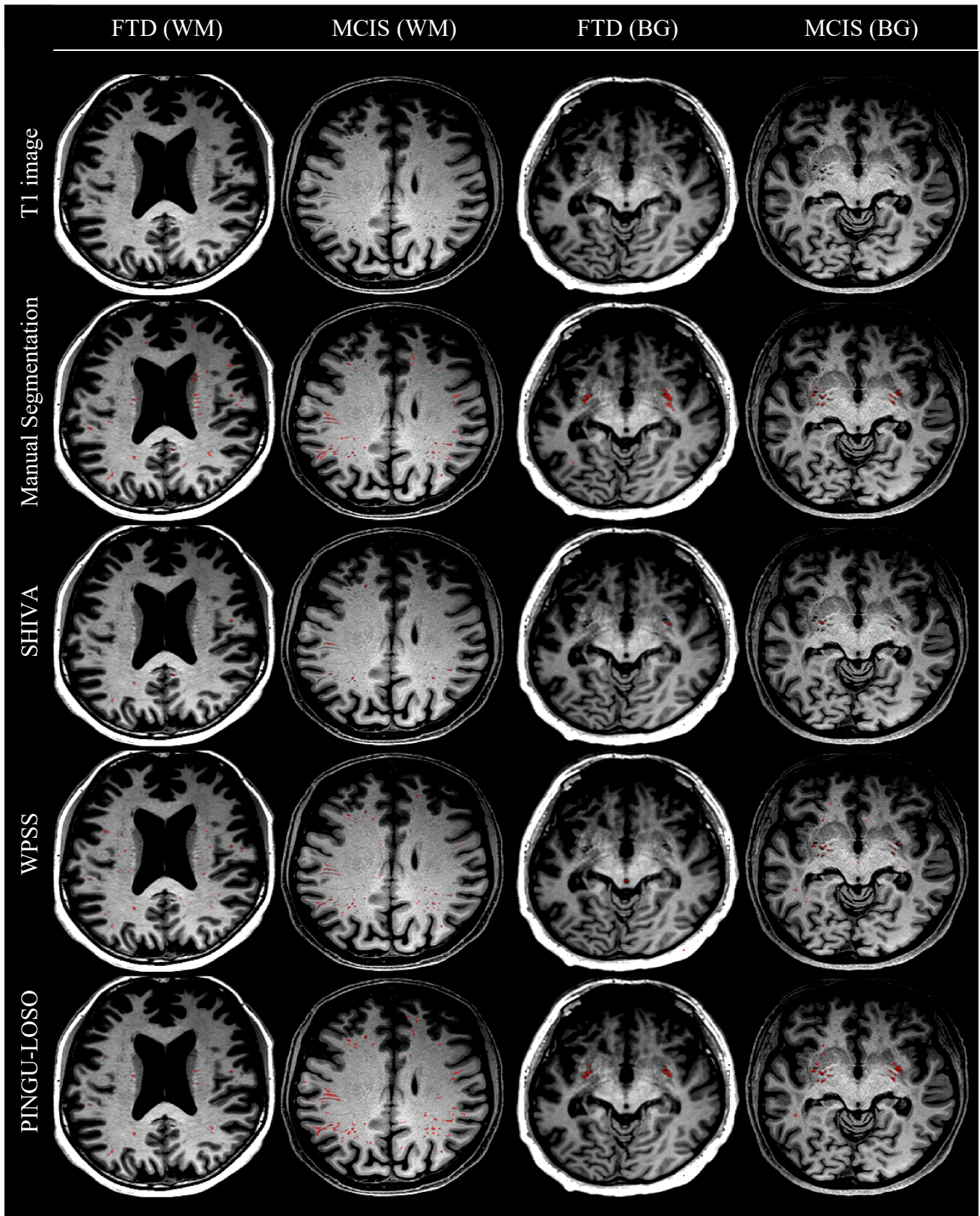

Figure 4: Overlay of the three deep learning algorithm segmentations on T1 images, and comparison to the manual segmentations, for one "high quality" dataset, MCIS, and one "low quality" dataset, FTD. Display slices were chosen automatically as those with the largest volume of PVS in WM and BG.

Table 5: Summary of average performance measures from external validation for each training + evaluation schedule. where mean is average over subjects for Dice Score, Sensitivity and Precision, and over datasets for Correlation. Highest performing algorithm for each measure indicated in bold font, PINGU-single site algorithms outperforming PINGU-LOSO indicated in italics.

| Region | Algorithm | Training Dataset | Evaluation Dataset | Dice Score voxel | Dice Score number | Sensitivity voxel | Sensitivity number | Precision voxel | Precision number | Correlation voxel | Correlation number |
|---|---|---|---|---|---|---|---|---|---|---|---|
| White Matter | SHIVA | External | All sites | 0.18 (0.13) | 0.29 (0.19) | 0.12 (0.10) | 0.21 (0.16) | 0.67 (0.19) | **0.77 (0.20)** | 0.80 (0.25) | 0.75 (0.24) |
| | WPSS | External | All sites | 0.30 (0.10) | 0.38 (0.18) | 0.27 (0.10) | **0.68 (0.21)** | 0.40 (0.17) | 0.28 (0.17) | 0.86 (0.04) | 0.49 (0.26) |
| | PINGU-LOSO | LOSO x 6 | Remaining 1 site x 6 | **0.38 (0.13)** | **0.58 (0.17)** | **0.34 (0.20)** | 0.50 (0.19) | **0.71 (0.23)** | 0.75 (0.18) | 0.82 (0.13) | 0.84 (0.15) |
| | PINGU-ADNI | ADNI | Remaining 5 sites | *0.39 (0.11)* | 0.50 (0.16) | 0.30 (0.11) | 0.36 (0.14) | 0.70 (0.18) | *0.81 (0.13)* | 0.84 (0.09) | 0.80 (0.14) |
| | PINGU-AF | AF | Remaining 5 sites | *0.47 (0.12)* | 0.57 (0.17) | *0.48 (0.17)* | *0.61 (0.22)* | 0.57 (0.19) | 0.59 (0.16) | 0.83 (0.21) | 0.82 (0.24) |
| | PINGU-CGS | CGS | Remaining 5 sites | 0.20 (0.09) | 0.42 (0.13) | 0.12 (0.06) | 0.27 (0.10) | *0.83 (0.13)* | *0.86 (0.12)* | 0.75 (0.13) | 0.81 (0.02) |
| | PINGU-MCIS | MCIS | Remaining 5 sites | 0.26 (0.11) | 0.52 (0.20) | 0.16 (0.08) | 0.38 (0.18) | *0.85 (0.09)* | *0.82 (0.11)* | 0.78 (0.07) | 0.75 (0.11) |
| Basal Ganglia | SHIVA | External | All 6 sites | 0.10 (0.11) | 0.15 (0.09) | 0.07 (0.09) | 0.09 (0.05) | 0.61 (0.28) | **0.79 (0.30)** | 0.11 (0.71) | 0.00 (0.40) |
| | WPSS | External | All 6 sites | 0.20 (0.09) | 0.37 (0.17) | 0.15 (0.10) | **0.56 (0.19)** | 0.39 (0.17) | 0.28 (0.15) | 0.56 (0.23) | 0.34 (0.15) |
| | PINGU-LOSO | LOSO x 6 | Remaining 1 site x 6 | **0.36 (0.13)** | **0.60 (0.20)** | **0.36 (0.26)** | 0.50 (0.17) | **0.70 (0.28)** | 0.76 (0.22) | 0.78 (0.12) | **0.46 (0.09)** |
| | PINGU-ADNI | ADNI | Remaining 5 sites | *0.46 (0.08)* | 0.55 (0.17) | *0.44 (0.14)* | 0.47 (0.16) | 0.57 (0.19) | 0.69 (0.18) | 0.79 (0.10) | *0.78 (0.28)* |
| | PINGU-AF | AF | Remaining 5 sites | *0.41 (0.11)* | 0.49 (0.15) | *0.46 (0.22)* | *0.51 (0.19)* | 0.53 (0.22) | 0.60 (0.20) | 0.18 (1.00) | *0.57 (0.14)* |
| | PINGU-CGS | CGS | Remaining 5 sites | 0.26 (0.16) | 0.46 (0.20) | 0.18 (0.14) | 0.30 (0.15) | 0.72 (0.28) | *0.80 (0.24)* | 0.52 (0.22) | *0.62 (0.21)* |
| | PINGU-MCIS | MCIS | Remaining 5 sites | 0.22 (0.09) | 0.55 (0.22) | 0.13 (0.06) | 0.36 (0.15) | *0.83 (0.25)* | *0.85 (0.18)* | 0.60 (0.01) | 0.45 (0.04) |

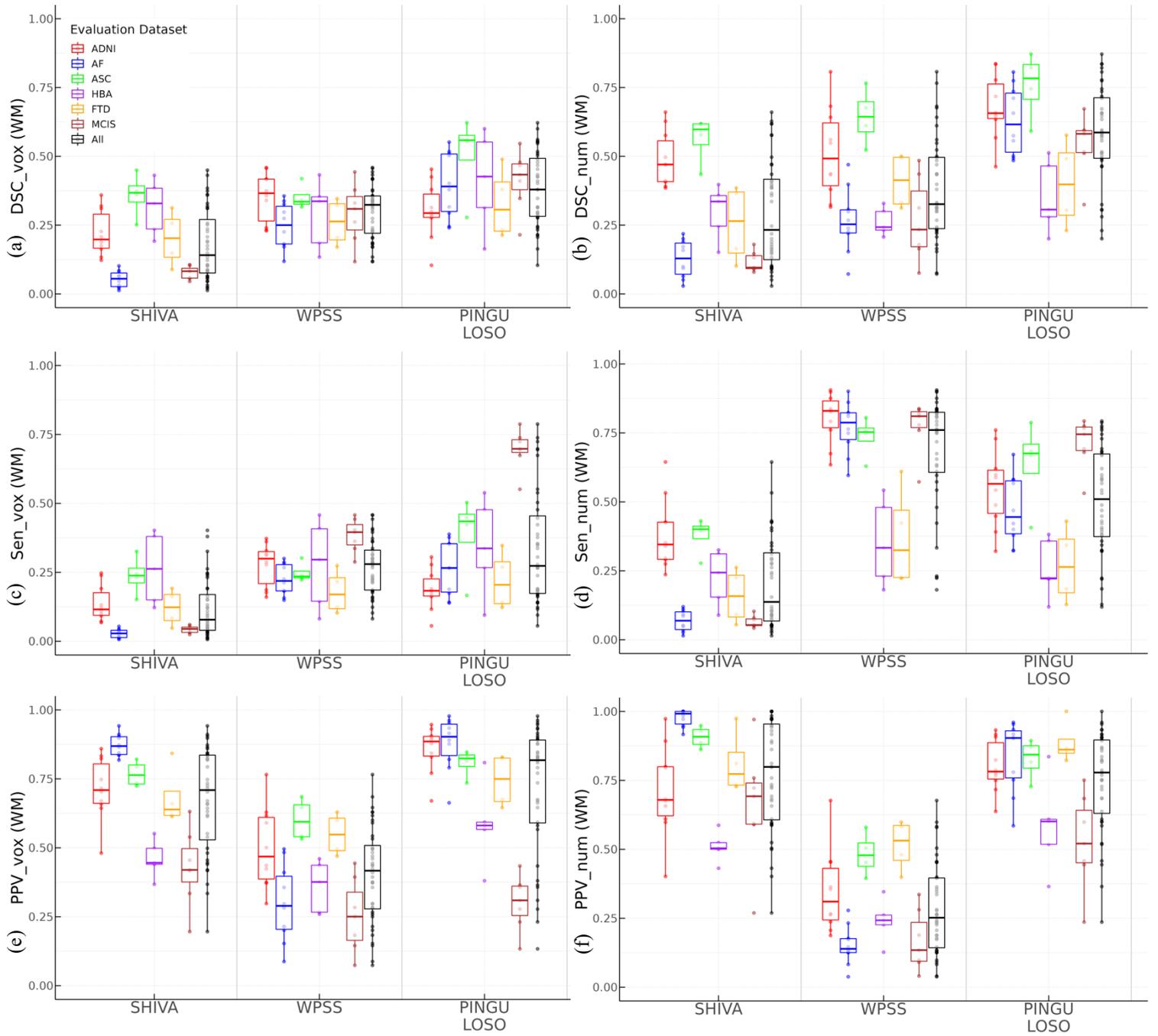

Figure 5: External validation, algorithm comparison (white matter (WM)): Performance of algorithms evaluated on unseen sites' data. DSC: dice score, Sen: sensitivity, PPV: Positive Predictive Value/Precision, vox: voxel-level metrics, num: cluster-level metrics.

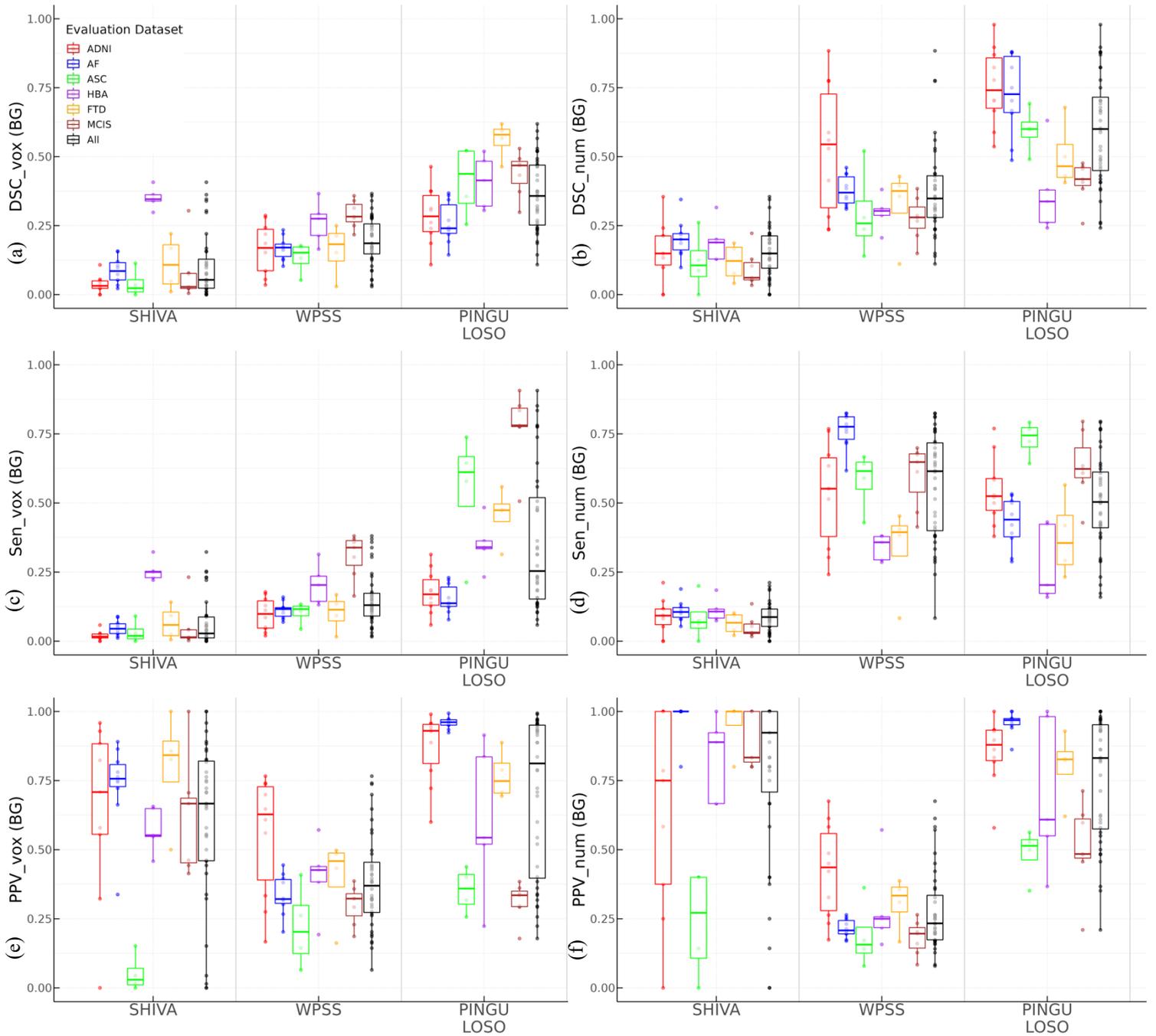

Figure 6: External validation, algorithm comparison (Basal Ganglia (BG)): Performance of algorithms evaluated on unseen sites' data. DSC: dice score, Sen: sensitivity, PPV: Positive Predictive Value/Precision, vox: voxel-level metrics, num: cluster-level metrics.

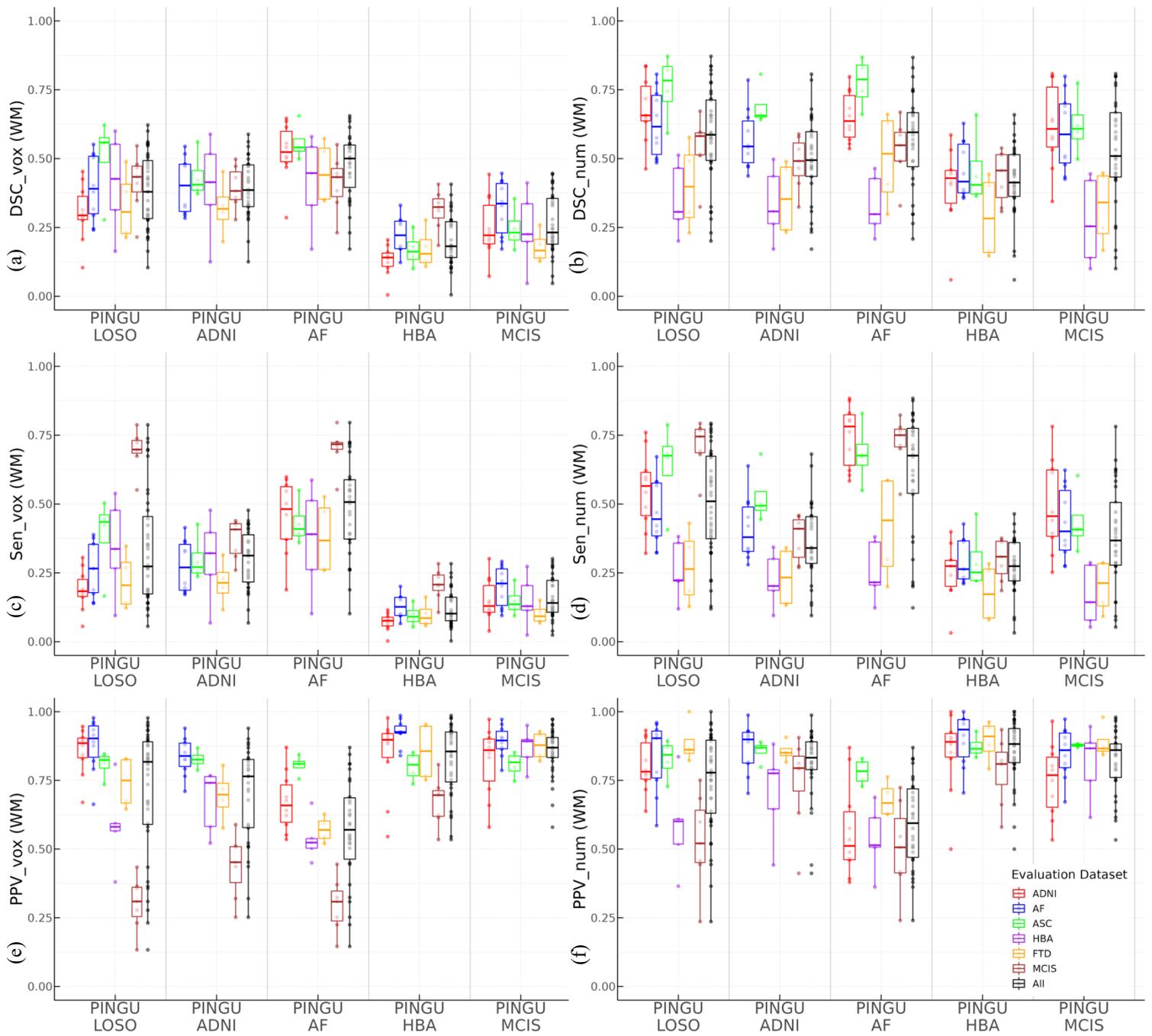

Figure 7: External validation, single site training (white matter (WM)): Performance for PINGU trained on single sites' data and evaluated on other sites' data, compared to PINGU trained and evaluated via leave one site out cross-validation. DSC: dice score, Sen: sensitivity, PPV: Positive Predictive Value/Precision, vox: voxel-level metrics, num: cluster-level metrics.

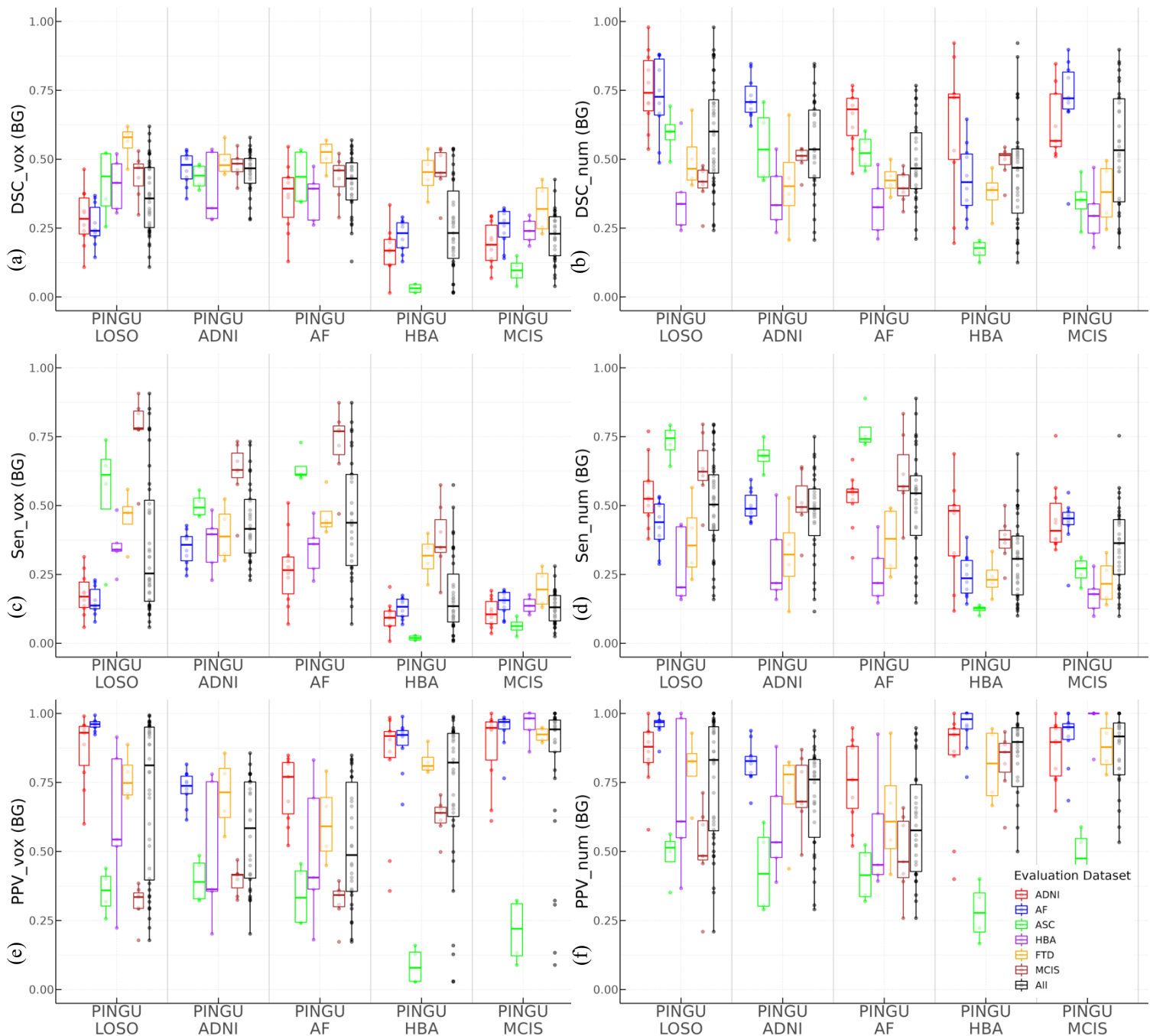

Figure 8: External validation, single site training (Basal Ganglia (BG)): Performance for PINGU trained on single sites' data and evaluated on other sites' data, compared to PINGU trained and evaluated via leave one site out cross-validation. DSC: dice score, Sen: sensitivity, PPV: Positive Predictive Value/Precision, vox: voxel-level metrics, num: cluster-level metrics.

*3.4 Methodological Variations*

We tested whether performance was improved by (1) using the most recent version of nnUNet (v2.2.1), or (2) allowing the nnUNet to automatically set the voxel spacing of input images, since the nnUNet documentation recommends comparing any variations to the base model. This was carried out in an internal 5FCV on all available data. nnUNet v2 resulted in slightly worse performance ($DSC_{vox}(WM)=0.46$, $DSC_{vox}(BG)=0.50$), and allowing nnUNet to set voxel spacing substantially reduced performance ($DSC_{vox}(WM)=0.39$, $DSC_{vox}(BG)=0.44$).

Table 6: Influence of nnUNet version and voxel spacing. Train Method = 5FCV, trained on all sites.

| Algorithm | Voxel Spacing | Dice score (WM) | Dice score (BG) |
| --- | --- | --- | --- |
| nnUNet v1 | 0.8x0.8x0.8 | 0.50 | 0.54 |
| nnUNet v2 | 0.8x0.8x0.8 | 0.46 | 0.50 |
| nnUNet v1 | Auto | 0.39 | 0.44 |
| nnUNet v2 | Auto | 0.29 | 0.34 |

4. Discussion

This study presents a nnUNet for segmentation of PVS on typical clinical quality T1w MRI images. The model had a dice score of 0.50 and 0.54 in the WM and Basal Ganglia respectively. The algorithm generalised better to unseen sites' MRI images than the other publicly available segmentation algorithms, especially in the Basal Ganglia, demonstrating superior generalisability. Contrary to expectations, training on a less diverse training data set sometimes yielded a model which performed better on unseen sites' MRI images.

Together the results indicate that PINGU finds around half of the manually labelled PVS ($Sen_{num}$=0.55-0.56), but less than half of the manually labelled voxels ($Sen_{vox}$=0.41-0.46). However, PINGU is moderately precise in measuring PVS voxels and clusters (PPV=0.74-0.77), and consequently the segmentations were strongly correlated (across subjects) with manual segmentations (r=0.69-0.84), thus accurately reflecting (with a scaling factor), enlarged PVS burden.

nnUNet is a leading algorithm in biomedical segmentation, giving the highest performance in 34 of 54 international biomedical segmentation challenges (Isensee et al., 2021), despite being a generalised algorithm not tailored to any specific segmentation task. While recent research explores the integration of vision foundation models (VFMs) and semi-supervised approaches inspired by multi-views and ensembling approaches to improve the accuracy of segmentation outcomes, such as Co-BioNet (Peiris et al., 2023), MC-Net+ (Wu et al., 2022), nnUNet remains as a state of the art model due to its systematic adaptation across all the phases including pre-processing, model configuration, training, inference, post-processing and ensembling. Notably, even with the rise of VFMs like MedSAM (Ma et al., 2024) and SAM-Med2D (Cheng et al., 2023), inspired by SAM (Kirillov et al., 2023), nnUNet maintains its superiority in the majority of biomedical segmentation tasks as these

VFM-inspired models struggle to segment medical images accurately with significant intra-variations such as tumor lesions and perivascular spaces. Additionally, nnUNet's robustness is reinforced by its extensive validation in real-world applications on various medical imaging modalities (i.e., CT, MRI, PET), where it continues to outperform competing methodologies. Its strength lies in the automatic selection of model parameters, in particular the interplay between patch size, batch size and model architecture, where patch size (the size of the subregion input to the network on each run), is iteratively reduced from the median full image size, and the model architecture altered accordingly, until the model can be trained with a batch size of at least 2 given GPU constraints, which is sufficient for most semantic segmentation tasks. This allows as much of the image to be input to the network as possible allowing maximal contextual information to be used.

The dice scores achieved with 5FCV trained on all-sites data (0.50 (WM), 0.54 (BG)), are in line with those reported in the literature (~0.25-0.75 Figure 5 of Waymont et al., 2024), though it should be noted that our evaluations include performance on data of varying demographics, field strength and resolutions (Table 1). For example, training and evaluating only on the site with the highest resolution and sample size, the Australian Football data, gave dice scores of 0.60 (WM) and 0.66 (BG), which is towards the higher end of the dice scores reported in the literature. This indicates that the nnUNet performed better when the training data was of high resolution, indeed the dataset with the second highest dice score on internal validation in both WM and BG was the MCIS dataset (Table 4), which was the other dataset with 0.8mm resolution.

External validation yielded substantially lower dice scores both for PINGU, and the two other publicly available deep learning algorithms. This highlights the fact that deep learning algorithms will often perform much worse on data with different characteristics to that which it was trained on. We attempted to overcome this by training on a diverse range of sites, with different scanner, image and cohort properties. The resulting algorithm had a dice score on external data of (0.38 (WM) and 0.36 (BG)). This was higher than SHIVA (0.18(WM 0.10(BG))) and WPSS (0.30(WM),0.20(BG)). In their respective publications, SHIVA reports dice scores of (0.55(WM), 0.74(BG)) and WPSS doesn't report a dice score, but a related measure balancing sensitivity and precision (F score) of 0.76 (whole brain). These reported performances are substantially higher than what we report using those algorithms, again highlights the poor generalisability of existing PVS segmentation algorithms to new data with different characteristics, and that dice scores reported in the literature should generally be considered 'best case scenarios'. Indeed, of the 8 DL PVS segmentation algorithms identified in Waymont et al. 2024, all of them trained and tested in data with the same scanner, image and cohort properties.

PINGU-LOSO may perform better on new sites' data compared to SHIVA and WPSS because it was trained on a wider range of sites with different characteristics, meaning the properties of new data will be more likely to have been learnt in the training phase. However, this assertion conflicts with the observation that greater performance on some measures (voxel-level dice score and sensitivity) on external sites was achieved by PINGU-AF and PINGU-ADNI, trained only on AF and ADNI data respectively, despite the far lower training set size (10 vs 30-36). One alternative explanation is that the nnUNet is inherently better than the UNet based networks trained in SHIVA and WPSS, due to its hyper-parameter selection procedure, yielding a superior training schedule. Alternatively, the nnUNet may have performed better in unseen sites' data due to the robust data-augmentation strategy employed, whereby input images are rotated, scaled, gaussian noised, gaussian-blurred, brightness-altered, contrast altered, simulated at low resolution, gamma corrected and mirrored. This exposes the network during training to a wide range of simulated image properties, thereby potentially increasing its generalisability to data with a wider range of characteristics. Finally, it could be simply due to the nature of the training data, or feature extraction. SHIVA trained solely on 1mm$^3$, whereas PINGU was trained with data up to 0.8mm$^3$. Since PVS are fine structures, a higher resolution could be crucial in examining and learning their structure. Indeed, the highest dice score reported of the 4 published 3D deep learning PVS segmentation algorithms was by (Lian et al., 2018), who trained and tested on 7T images with resolution up to 0.4x0.4x0.4mm$^3$. WPSS on the other hand had a very different training strategy, where in addition to the raw image, a Frangi filtered image highlighting vessel-like structures (Frangi et al., 1998) was input as an additional channel to the network. In our data WPSS had a relatively high sensitivity and relatively low precision. This could indicate that the Frangi filter highlights a lot more than just PVS, and other structures and/or noise are detected as PVS, or alternatively that the Frangi filter has highlighted true PVS which were not observed by our manual raters.

Despite the moderate to low voxel-level dice scores across algorithms and training schedules, the correlations with manually segmented PVS volume were generally high, and the cluster level dice scores, sensitivities, precisions and correlations were almost always higher than the voxel level performance metrics. This indicates that PINGU detects a large proportion of the PVS, with moderate precision, but does not segment each individual PVS in its entirety, to the same extent as the manual raters. This observation could favour the usage of cluster/count level measures of PVS over volume, or it could bolster the confidence in the measure of volume, as it linearly and strongly relates to the true volume, as quantified in the voxel-level correlations.

The improved performance in the basal ganglia (both compared to in the white matter and compared to other algorithms) is noteworthy, since an emerging pattern in the literature is that PVS in the basal ganglia are more related to vascular pathologies than PVS in the WM (Rowsthorn et al., 2023; Smeijer et al., 2019; Tu et al., 2022; Yamasaki et al., 2022), whereas

WM PVS are more related to amyloid-related pathologies (Charidimou et al., 2017; Wang et al., 2022). Thus PINGU may be particularly useful in investigating glymphatics in neurovascular pathologies (cerebral small vessel disease, vascular dementia, stroke, hypertension etc.).

The first limitation of this study is the inherent difficulty and uncertainty in manually segmenting PVS. Since PVS are small, fine structures, often indistinguishable from noise, and similar in appearance to other structures, identification is often challenging. Consequently, although inter-rater reliability of PVS manual segmentation has not been reported in the literature, it is likely low, and places an upper limit on the dice scores of the algorithms. Hence, algorithms acknowledging and accounting for noisy labels (Peiris et al., 2023; Wang et al., 2021) would likely be useful for this task. Secondly, each dataset used had a variable number of subjects, meaning our performance measures of algorithms trained on different datasets are not directly comparable, and may reflect training dataset size. Finally, although the sample size we have used is typical in comparison to other deep learning algorithms for 3D PVS segmentation, we have also intentionally incorporated more heterogeneity. The more variance in a dataset, the larger the sample required to model that variance, and hence we may be underpowered to achieve optimal modelling of the variance in our dataset.

PINGU also has several strengths. Firstly, it produces a full 3D segmentation of PVS, which is only the case for a minority of PVS detection algorithms (Table 3) due to the time required to manually segment PVS in 3D. Importantly, PINGU performs favourably on unseen sites' data compared to other publicly available 3D PVS segmentation algorithms (Table 4), advocating for its superior utility on new sites' data. Next, PINGU is more optimised for use on clinical scans compared to existing algorithms for two reasons: PINGU only requires a 3D T1w image, which is the most commonly clinically, and research acquired modality, in contrast to other algorithms which require T1w + T2w ((Lan et al., 2023), EPC version), T2w (Lian et al., 2018); and secondly, PINGU was trained and tested on typical clinical resolutions of 0.8-1.2mm, and typical clinical field strengths (1.5T-3T), making it appropriate for generalised usage on clinical scans. PINGU was trained on manual segmentations with BG labels, and offers improved BG segmentation, a region often missed by other models, and a region where the PVSs are related to vascular disease and pathology. A strength of this study (as opposed to of the model) is that models were evaluated on hold-out data sets from sites that had not been used in the training phase. This practice is strongly encouraged in deep learning for neuroimaging to demonstrate the generalisability of machine learning algorithms, but to our knowledge has not been done previously for the task of 3D PVS segmentation.

Supplementary Material

Table S1: Dice score, voxel-level, for each algorithm on each dataset. LO: Leave Out, 5FCV: Five fold cross validation. LO: Leave Out. Average (standard deviation) is mean across subjects in validation set(s).

| Region | Algorithm | Training Dataset | Internal Validation (5FCV) | External Validation | | | | | | |
|---|---|---|---|---|---|---|---|---|---|---|
| | | | | ADNI | AF | ASC | HBA | FTD | MCIS | Average |
| White Matter | SHIVA | - | - | 0.22 (0.08) | 0.05 (0.03) | 0.36 (0.08) | 0.31 (0.10) | 0.20 (0.10) | 0.08 (0.02) | 0.18 (0.13) |
| | WPSS | - | - | 0.35 (0.09) | 0.25 (0.08) | 0.35 (0.05) | 0.29 (0.12) | 0.26 (0.09) | 0.29 (0.11) | 0.30 (0.10) |
| | PINGU | All Sites | 0.50 (0.15) | - | - | - | - | - | - | - |
| | PINGU | LO ADNI | 0.50 (0.14) | 0.30 (0.10) | - | - | - | - | - | 0.38 (0.13) |
| | | LO AF | 0.46 (0.13) | - | 0.40 (0.12) | - | - | - | - | |
| | | LO ASC | 0.48 (0.15) | - | - | 0.50 (0.15) | - | - | - | |
| | | LO HBA | 0.53 (0.12) | - | - | - | 0.41 (0.18) | - | - | |
| | | LO FTD | 0.52 (0.13) | - | - | - | - | 0.33 (0.13) | - | |
| | | LO MCIS | 0.49 (0.15) | - | - | - | - | - | 0.41 (0.11) | |
| | PINGU | ADNI | 0.45 (0.11) | - | 0.40 (0.10) | 0.44 (0.09) | 0.40 (0.18) | 0.32 (0.10) | 0.39 (0.08) | - |
| | | AF | 0.60 (0.08) | 0.52 (0.10) | - | 0.56 (0.07) | 0.41 (0.17) | 0.45 (0.12) | 0.41 (0.10) | 0.52 (0.10) |
| | | HBA | 0.23 (0.13) | 0.13 (0.06) | 0.22 (0.07) | 0.17 (0.06) | - | 0.17 (0.08) | 0.31 (0.07) | 0.13 (0.06) |
| | | MCIS | 0.51 (0.06) | 0.25 (0.11) | 0.32 (0.10) | 0.25 (0.08) | 0.24 (0.14) | 0.18 (0.06) | - | 0.25 (0.11) |
| Basal Ganglia | SHIVA | - | - | 0.04 (0.03) | 0.09 (0.05) | 0.04 (0.05) | 0.35 (0.04) | 0.11 (0.10) | 0.08 (0.10) | 0.10 (0.11) |
| | WPSS | - | - | 0.16 (0.09) | 0.17 (0.04) | 0.13 (0.06) | 0.26 (0.08) | 0.16 (0.10) | 0.29 (0.05) | 0.20 (0.09) |
| | PINGU | All Sites | 0.54 (0.11) | - | - | - | - | - | - | - |
| | PINGU | LO ADNI | 0.56 (0.11) | 0.29 (0.10) | - | - | - | - | - | 0.36 (0.13) |
| | | LO AF | 0.48 (0.12) | - | 0.26 (0.07) | - | - | - | - | |
| | | LO ASC | 0.53 (0.11) | - | - | 0.41 (0.13) | - | - | - | |
| | | LO HBA | 0.56 (0.10) | - | - | - | 0.41 (0.10) | - | - | |
| | | LO FTD | 0.54 (0.12) | - | - | - | - | 0.56 (0.07) | - | |
| | | LO MCIS | 0.54 (0.13) | - | - | - | - | - | 0.44 (0.08) | |
| | PINGU | ADNI | 0.50 (0.08) | - | 0.47 (0.06) | 0.44 (0.05) | 0.39 (0.13) | 0.50 (0.06) | 0.48 (0.05) | 0.46 (0.08) |
| | | AF | 0.66 (0.03) | 0.37 (0.12) | - | 0.44 (0.10) | 0.36 (0.09) | 0.52 (0.06) | 0.43 (0.08) | 0.41 (0.11) |
| | | HBA | 0.28 (0.10) | 0.17 (0.09) | 0.22 (0.06) | 0.03 (0.02) | - | 0.45 (0.08) | 0.46 (0.09) | 0.26 (0.16) |
| | | MCIS | 0.56 (0.07) | 0.19 (0.08) | 0.25 (0.07) | 0.10 (0.05) | 0.24 (0.05) | 0.32 (0.10) | - | 0.22 (0.09) |

Table S2: Dice score, cluster-level, for each algorithm on each dataset. LO: Leave Out, 5FCV: Five fold cross validation. LO: Leave Out. Average (standard deviation) is mean across subjects in validation set(s).

| Region | Algorithm | Training Dataset | Internal Validation (5FCV) | External Validation ADNI | AF | ASC | HBA | FTD | MCIS | Average |
|---|---|---|---|---|---|---|---|---|---|---|
| White Matter | SHIVA | - | - | 0.49 (0.10) | 0.13 (0.07) | 0.56 (0.09) | 0.30 (0.10) | 0.25 (0.14) | 0.12 (0.04) | 0.29 (0.19) |
| | WPSS | - | - | 0.51 (0.16) | 0.27 (0.11) | 0.64 (0.10) | 0.26 (0.05) | 0.41 (0.10) | 0.27 (0.15) | 0.38 (0.18) |
| | PINGU | All Sites | 0.63 (0.17) | - | - | - | - | - | - | - |
| | PINGU | LO ADNI | 0.60 (0.18) | 0.68 (0.12) | - | - | - | - | - | 0.58 (0.17) |
| | | LO AF | 0.62 (0.19) | - | 0.63 (0.12) | - | - | - | - | |
| | | LO ASC | 0.61 (0.18) | - | - | 0.76 (0.12) | - | - | - | |
| | | LO HBA | 0.67 (0.14) | - | - | - | 0.35 (0.13) | - | - | |
| | | LO FTD | 0.65 (0.17) | - | - | - | - | 0.40 (0.16) | - | |
| | | LO MCIS | 0.62 (0.18) | - | - | - | - | - | 0.54 (0.11) | |
| | PINGU | ADNI | 0.70 (0.09) | - | 0.57 (0.11) | 0.69 (0.08) | 0.34 (0.13) | 0.36 (0.14) | 0.48 (0.09) | 0.50 (0.16) |
| | | AF | 0.65 (0.09) | 0.65 (0.09) | - | 0.78 (0.09) | 0.33 (0.11) | 0.50 (0.18) | 0.53 (0.11) | 0.57 (0.17) |
| | | HBA | 0.25 (0.11) | 0.40 (0.15) | 0.46 (0.10) | 0.46 (0.14) | - | 0.29 (0.16) | 0.44 (0.10) | 0.42 (0.13) |
| | | MCIS | 0.66 (0.09) | 0.62 (0.15) | 0.60 (0.14) | 0.62 (0.11) | 0.27 (0.16) | 0.32 (0.14) | - | 0.52 (0.20) |
| Basal Ganglia | SHIVA | - | - | 0.16 (0.11) | 0.20 (0.07) | 0.12 (0.11) | 0.19 (0.08) | 0.12 (0.07) | 0.09 (0.07) | 0.15 (0.09) |
| | WPSS | - | - | 0.53 (0.23) | 0.38 (0.05) | 0.29 (0.16) | 0.30 (0.06) | 0.32 (0.14) | 0.28 (0.08) | 0.37 (0.17) |
| | PINGU | All Sites | 0.66 (0.17) | - | - | - | - | - | - | - |
| | PINGU | LO ADNI | 0.59 (0.16) | 0.75 (0.14) | - | - | - | - | - | 0.60 (0.20) |
| | | LO AF | 0.63 (0.18) | - | 0.72 (0.14) | - | - | - | - | |
| | | LO ASC | 0.64 (0.19) | - | - | 0.60 (0.08) | - | - | - | |
| | | LO HBA | 0.67 (0.15) | - | - | - | 0.37 (0.16) | - | - | |
| | | LO FTD | 0.65 (0.15) | - | - | - | - | 0.50 (0.12) | - | |
| | | LO MCIS | 0.66 (0.19) | - | - | - | - | - | 0.41 (0.07) | |
| | PINGU | ADNI | 0.77 (0.10) | - | 0.72 (0.08) | 0.55 (0.14) | 0.36 (0.12) | 0.42 (0.19) | 0.50 (0.05) | 0.55 (0.17) |
| | | AF | 0.67 (0.09) | 0.65 (0.10) | - | 0.53 (0.07) | 0.33 (0.11) | 0.43 (0.06) | 0.40 (0.06) | 0.49 (0.15) |
| | | HBA | 0.33 (0.11) | 0.62 (0.25) | 0.43 (0.13) | 0.17 (0.04) | - | 0.38 (0.08) | 0.49 (0.06) | 0.46 (0.20) |
| | | MCIS | 0.63 (0.08) | 0.67 (0.17) | 0.72 (0.16) | 0.35 (0.09) | 0.30 (0.11) | 0.38 (0.12) | - | 0.55 (0.22) |

Table S3: Recall (Sensitivity), voxel level, for each algorithm on each dataset. LO: Leave Out, 5FCV: Five fold cross validation. LOSOCV: Leave Out One Site Cross Validation. Average (standard deviation) is mean across subjects in validation set(s).

| Region | Algorithm | Training Dataset | Internal Validation (5FCV) | External Validation | | | | | | |
|---|---|---|---|---|---|---|---|---|---|---|
| | | | | ADNI | AF | ASC | HBA | FTD | MCIS | Average |
| White Matter | SHIVA | - | - | 0.14 (0.07) | 0.03 (0.02) | 0.24 (0.07) | 0.26 (0.13) | 0.12 (0.07) | 0.04 (0.01) | 0.12 (0.10) |
| | WPSS | - | - | 0.28 (0.08) | 0.22 (0.06) | 0.25 (0.04) | 0.28 (0.16) | 0.18 (0.08) | 0.38 (0.06) | 0.27 (0.10) |
| | PINGU | All Sites | 0.41 (0.16) | - | - | - | - | - | - | - |
| | PINGU | LO ADNI | 0.43 (0.17) | 0.19 (0.07) | - | - | - | - | - | 0.34 (0.20) |
| | | LO AF | 0.35 (0.13) | - | 0.26 (0.10) | - | - | - | - | |
| | | LO ASC | 0.39 (0.17) | - | - | 0.38 (0.15) | - | - | - | |
| | | LO HBA | 0.44 (0.15) | - | - | - | 0.34 (0.18) | - | - | |
| | | LO FTD | 0.44 (0.15) | - | - | - | - | 0.22 (0.11) | - | |
| | | LO MCIS | 0.39 (0.17) | - | - | - | - | - | 0.70 (0.07) | |
| | PINGU | ADNI | 0.33 (0.10) | - | 0.27 (0.09) | 0.30 (0.09) | 0.30 (0.16) | 0.21 (0.08) | 0.37 (0.07) | 0.30 (0.11) |
| | | AF | 0.54 (0.13) | 0.45 (0.14) | - | 0.43 (0.08) | 0.37 (0.19) | 0.38 (0.14) | 0.70 (0.07) | 0.48 (0.17) |
| | | HBA | 0.14 (0.08) | 0.07 (0.03) | 0.13 (0.04) | 0.10 (0.04) | - | 0.10 (0.05) | 0.21 (0.06) | 0.12 (0.06) |
| | | MCIS | 0.42 (0.08) | 0.15 (0.08) | 0.20 (0.07) | 0.15 (0.06) | 0.15 (0.09) | 0.10 (0.04) | - | 0.16 (0.08) |
| Basal Ganglia | SHIVA | - | - | 0.02 (0.02) | 0.05 (0.03) | 0.03 (0.04) | 0.26 (0.04) | 0.07 (0.06) | 0.05 (0.08) | 0.07 (0.09) |
| | WPSS | - | - | 0.10 (0.06) | 0.11 (0.03) | 0.10 (0.04) | 0.21 (0.07) | 0.10 (0.07) | 0.31 (0.08) | 0.15 (0.10) |
| | PINGU | All Sites | 0.46 (0.15) | - | - | - | - | - | - | - |
| | PINGU | LO ADNI | 0.50 (0.16) | 0.18 (0.08) | - | - | - | - | - | 0.36 (0.26) |
| | | LO AF | 0.38 (0.15) | - | 0.15 (0.05) | - | - | - | - | |
| | | LO ASC | 0.46 (0.16) | - | - | 0.54 (0.23) | - | - | - | |
| | | LO HBA | 0.50 (0.15) | - | - | - | 0.35 (0.09) | - | - | |
| | | LO FTD | 0.48 (0.16) | - | - | - | - | 0.45 (0.10) | - | |
| | | LO MCIS | 0.45 (0.16) | - | - | - | - | - | 0.78 (0.13) | |
| | PINGU | ADNI | 0.39 (0.10) | - | 0.35 (0.06) | 0.50 (0.04) | 0.36 (0.10) | 0.40 (0.11) | 0.62 (0.11) | 0.44 (0.14) |
| | | AF | 0.61 (0.05) | 0.27 (0.13) | - | 0.64 (0.06) | 0.34 (0.10) | 0.47 (0.08) | 0.72 (0.13) | 0.46 (0.22) |
| | | HBA | 0.18 (0.08) | 0.10 (0.05) | 0.13 (0.04) | 0.02 (0.01) | - | 0.31 (0.08) | 0.38 (0.13) | 0.18 (0.14) |
| | | MCIS | 0.49 (0.12) | 0.11 (0.05) | 0.15 (0.04) | 0.06 (0.03) | 0.14 (0.03) | 0.20 (0.07) | - | 0.13 (0.06) |

Table S4: Cluster Recall (Sensitivity), cluster level , for each algorithm on each dataset. LO: Leave Out, 5FCV: Five fold cross validation. LOSOCV: Leave Out One Site Cross Validation. Average (standard deviation) is mean across subjects in validation set(s).

| Region | Algorithm | Training Dataset | Internal Validation (5FCV) | External Validation | | | | | | |
|---|---|---|---|---|---|---|---|---|---|---|
| | | | | ADNI | AF | ASC | HBA | FTD | MCIS | Average |
| White Matter | SHIVA | - | - | 0.38 (0.13) | 0.07 (0.04) | 0.38 (0.07) | 0.22 (0.10) | 0.16 (0.10) | 0.06 (0.02) | 0.21 (0.16) |
| | WPSS | - | - | 0.80 (0.09) | 0.77 (0.09) | 0.73 (0.07) | 0.35 (0.16) | 0.37 (0.18) | 0.77 (0.09) | 0.68 (0.21) |
| | PINGU | All Sites | 0.55 (0.20) | - | - | - | - | - | - | - |
| | PINGU | LO ADNI | 0.54 (0.21) | 0.55 (0.14) | - | - | - | - | - | 0.50 (0.19) |
| | | LO AF | 0.52 (0.20) | - | 0.47 (0.12) | - | - | - | - | |
| | | LO ASC | 0.52 (0.21) | - | - | 0.64 (0.16) | - | - | - | |
| | | LO HBA | 0.62 (0.17) | - | - | - | 0.26 (0.11) | - | - | |
| | | LO FTD | 0.58 (0.19) | - | - | - | - | 0.27 (0.14) | - | |
| | | LO MCIS | 0.54 (0.21) | - | - | - | - | - | 0.71 (0.09) | |
| | PINGU | ADNI | 0.60 (0.12) | - | 0.41 (0.11) | 0.53 (0.10) | 0.23 (0.10) | 0.23 (0.11) | 0.38 (0.08) | 0.36 (0.14) |
| | | AF | 0.65 (0.12) | 0.75 (0.11) | - | 0.68 (0.11) | 0.26 (0.11) | 0.42 (0.20) | 0.72 (0.09) | 0.61 (0.22) |
| | | HBA | 0.15 (0.07) | 0.25 (0.10) | 0.29 (0.08) | 0.30 (0.11) | - | 0.18 (0.11) | 0.30 (0.08) | 0.27 (0.10) |
| | | MCIS | 0.58 (0.08) | 0.49 (0.17) | 0.43 (0.13) | 0.44 (0.12) | 0.17 (0.11) | 0.20 (0.10) | - | 0.38 (0.18) |
| Basal Ganglia | SHIVA | - | - | 0.09 (0.06) | 0.11 (0.04) | 0.08 (0.08) | 0.11 (0.04) | 0.06 (0.04) | 0.05 (0.04) | 0.09 (0.05) |
| | WPSS | - | - | 0.53 (0.19) | 0.76 (0.06) | 0.58 (0.11) | 0.34 (0.05) | 0.33 (0.17) | 0.60 (0.11) | 0.56 (0.19) |
| | PINGU | All Sites | 0.56 (0.17) | - | - | - | - | - | - | - |
| | PINGU | LO ADNI | 0.53 (0.18) | 0.55 (0.12) | - | - | - | - | - | 0.47 (0.16) |
| | | LO AF | 0.51 (0.17) | - | 0.43 (0.09) | - | - | - | - | |
| | | LO ASC | 0.55 (0.17) | - | - | 0.73 (0.07) | - | - | - | |
| | | LO HBA | 0.60 (0.14) | - | - | - | 0.28 (0.14) | - | - | |
| | | LO FTD | 0.58 (0.15) | - | - | - | - | 0.38 (0.15) | - | |
| | | LO MCIS | 0.56 (0.18) | - | - | - | - | - | 0.63 (0.12) | |
| | PINGU | ADNI | 0.66 (0.06) | - | 0.50 (0.05) | 0.68 (0.06) | 0.30 (0.16) | 0.32 (0.17) | 0.51 (0.11) | 0.47 (0.16) |
| | | AF | 0.63 (0.10) | 0.52 (0.10) | - | 0.77 (0.08) | 0.25 (0.11) | 0.37 (0.13) | 0.61 (0.15) | 0.51 (0.19) |
| | | HBA | 0.22 (0.12) | 0.41 (0.18) | 0.24 (0.08) | 0.12 (0.02) | - | 0.24 (0.07) | 0.37 (0.08) | 0.30 (0.15) |
| | | MCIS | 0.57 (0.12) | 0.46 (0.13) | 0.44 (0.09) | 0.26 (0.05) | 0.18 (0.07) | 0.23 (0.09) | - | 0.36 (0.15) |

Table S5: Precision (Positive Predictive Value), for each algorithm on each dataset. LO: Leave Out, 5FCV: Five fold cross validation. LOSOCV: Leave Out One Site Cross Validation. Average (standard deviation) is mean across subjects in validation set(s).

| Region | Algorithm | Training Dataset | Internal Validation (5FCV) | External Validation ADNI | AF | ASC | HBA | FTD | MCIS | Average |
|---|---|---|---|---|---|---|---|---|---|---|
| White Matter | SHIVA | - | - | 0.71 (0.11) | 0.87 (0.04) | 0.77 (0.05) | 0.46 (0.07) | 0.68 (0.11) | 0.43 (0.14) | 0.67 (0.19) |
| | WPSS | - | - | 0.50 (0.15) | 0.30 (0.14) | 0.60 (0.08) | 0.36 (0.09) | 0.55 (0.08) | 0.25 (0.13) | 0.40 (0.17) |
| | PINGU | All Sites | 0.75 (0.14) | - | - | - | - | - | - | - |
| | PINGU | LO ADNI | 0.71 (0.17) | 0.86 (0.08) | - | - | - | - | - | 0.71 (0.23) |
| | | LO AF | 0.77 (0.11) | - | 0.88 (0.10) | - | - | - | - | |
| | | LO ASC | 0.74 (0.13) | - | - | 0.81 (0.05) | - | - | - | |
| | | LO HBA | 0.73 (0.13) | - | - | - | 0.59 (0.15) | - | - | |
| | | LO FTD | 0.74 (0.14) | - | - | - | - | 0.74 (0.10) | - | |
| | | LO MCIS | 0.77 (0.11) | - | - | - | - | - | 0.30 (0.10) | |
| | PINGU | ADNI | 0.80 (0.07) | - | 0.84 (0.07) | 0.83 (0.03) | 0.68 (0.11) | 0.69 (0.09) | 0.44 (0.12) | 0.70 (0.18) |
| | | AF | 0.73 (0.15) | 0.67 (0.11) | - | 0.81 (0.04) | 0.54 (0.08) | 0.57 (0.05) | 0.30 (0.10) | 0.57 (0.19) |
| | | HBA | 0.81 (0.14) | 0.84 (0.14) | 0.92 (0.05) | 0.80 (0.05) | - | 0.85 (0.11) | 0.67 (0.09) | 0.83 (0.13) |
| | | MCIS | 0.68 (0.08) | 0.82 (0.13) | 0.89 (0.06) | 0.81 (0.05) | 0.87 (0.07) | 0.88 (0.05) | - | 0.85 (0.09) |
| Basal Ganglia | SHIVA | - | - | 0.64 (0.32) | 0.73 (0.15) | 0.05 (0.07) | 0.57 (0.08) | 0.80 (0.21) | 0.62 (0.21) | 0.61 (0.28) |
| | WPSS | - | - | 0.55 (0.22) | 0.34 (0.07) | 0.22 (0.15) | 0.40 (0.14) | 0.39 (0.16) | 0.30 (0.07) | 0.39 (0.17) |
| | PINGU | All Sites | 0.74 (0.11) | - | - | - | - | - | - | - |
| | PINGU | LO ADNI | 0.70 (0.13) | 0.87 (0.13) | - | - | - | - | - | 0.70 (0.28) |
| | | LO AF | 0.76 (0.11) | - | 0.96 (0.02) | - | - | - | - | |
| | | LO ASC | 0.72 (0.14) | - | - | 0.35 (0.08) | - | - | - | |
| | | LO HBA | 0.71 (0.13) | - | - | - | 0.61 (0.28) | - | - | |
| | | LO FTD | 0.70 (0.13) | - | - | - | - | 0.77 (0.09) | - | |
| | | LO MCIS | 0.76 (0.13) | - | - | - | - | - | 0.31 (0.07) | |
| | PINGU | ADNI | 0.76 (0.08) | - | 0.73 (0.06) | 0.40 (0.08) | 0.49 (0.26) | 0.71 (0.14) | 0.40 (0.05) | 0.57 (0.19) |
| | | AF | 0.73 (0.08) | 0.73 (0.12) | - | 0.34 (0.11) | 0.49 (0.26) | 0.61 (0.15) | 0.32 (0.07) | 0.53 (0.22) |
| | | HBA | 0.80 (0.14) | 0.82 (0.22) | 0.89 (0.09) | 0.09 (0.07) | - | 0.83 (0.05) | 0.62 (0.07) | 0.72 (0.28) |
| | | MCIS | 0.68 (0.10) | 0.88 (0.14) | 0.94 (0.07) | 0.21 (0.12) | 0.96 (0.06) | 0.92 (0.03) | - | 0.83 (0.25) |

Table S6: Cluster Precision/PPV (Positive Predictive Value), for each algorithm on each dataset. LO: Leave Out, 5FCV: Five fold cross validation. LOSOCV: Leave Out One Site Cross Validation. Average (standard deviation) is mean across subjects in validation set(s).

| Region | Algorithm | Training Dataset | Internal Validation (5FCV) | External Validation ADNI | AF | ASC | HBA | FTD | MCIS | Average |
|---|---|---|---|---|---|---|---|---|---|---|
| White Matter | SHIVA | - | - | 0.71 (0.16) | 0.98 (0.03) | 0.91 (0.04) | 0.51 (0.06) | 0.81 (0.11) | 0.66 (0.21) | 0.77 (0.20) |
| | WPSS | - | - | 0.35 (0.15) | 0.15 (0.07) | 0.48 (0.08) | 0.24 (0.08) | 0.52 (0.09) | 0.17 (0.11) | 0.28 (0.17) |
| | PINGU | All Sites | 0.77 (0.14) | - | - | - | - | - | - | - |
| | PINGU | LO ADNI | 0.73 (0.17) | 0.80 (0.09) | - | - | - | - | - | 0.75 (0.18) |
| | | LO AF | 0.79 (0.12) | - | 0.84 (0.13) | - | - | - | - | |
| | | LO ASC | 0.76 (0.13) | - | - | 0.83 (0.07) | - | - | - | |
| | | LO HBA | 0.74 (0.14) | - | - | - | 0.59 (0.17) | - | - | |
| | | LO FTD | 0.75 (0.15) | - | - | - | - | 0.89 (0.08) | - | |
| | | LO MCIS | 0.77 (0.14) | - | - | - | - | - | 0.53 (0.17) | |
| | PINGU | ADNI | 0.79 (0.10) | - | 0.87 (0.09) | 0.86 (0.04) | 0.71 (0.17) | 0.86 (0.04) | 0.74 (0.17) | 0.81 (0.13) |
| | | AF | 0.68 (0.20) | 0.56 (0.17) | - | 0.78 (0.05) | 0.54 (0.12) | 0.68 (0.07) | 0.50 (0.17) | 0.59 (0.16) |
| | | HBA | 0.80 (0.21) | 0.85 (0.15) | 0.90 (0.10) | 0.87 (0.04) | - | 0.89 (0.08) | 0.79 (0.12) | 0.86 (0.12) |
| | | MCIS | 0.75 (0.14) | 0.75 (0.13) | 0.85 (0.09) | 0.88 (0.01) | 0.81 (0.13) | 0.89 (0.06) | - | 0.82 (0.11) |
| Basal Ganglia | SHIVA | - | - | 0.64 (0.36) | 0.98 (0.06) | 0.24 (0.20) | 0.83 (0.15) | 0.95 (0.10) | 0.90 (0.10) | 0.79 (0.30) |
| | WPSS | - | - | 0.42 (0.17) | 0.21 (0.03) | 0.19 (0.12) | 0.29 (0.16) | 0.31 (0.10) | 0.18 (0.06) | 0.28 (0.15) |
| | PINGU | All Sites | 0.76 (0.13) | - | - | - | - | - | - | - |
| | PINGU | LO ADNI | 0.70 (0.15) | 0.86 (0.12) | - | - | - | - | - | 0.76 (0.22) |
| | | LO AF | 0.79 (0.12) | - | 0.96 (0.04) | - | - | - | - | |
| | | LO ASC | 0.75 (0.16) | - | - | 0.49 (0.09) | - | - | - | |
| | | LO HBA | 0.71 (0.14) | - | - | - | 0.70 (0.28) | - | - | |
| | | LO FTD | 0.71 (0.14) | - | - | - | - | 0.80 (0.13) | - | |
| | | LO MCIS | 0.75 (0.16) | - | - | - | - | - | 0.51 (0.16) | |
| | PINGU | ADNI | 0.74 (0.09) | - | 0.82 (0.08) | 0.43 (0.16) | 0.60 (0.20) | 0.70 (0.18) | 0.71 (0.13) | 0.69 (0.18) |
| | | AF | 0.62 (0.11) | 0.75 (0.15) | - | 0.42 (0.10) | 0.56 (0.22) | 0.64 (0.22) | 0.49 (0.15) | 0.60 (0.20) |
| | | HBA | 0.84 (0.15) | 0.83 (0.21) | 0.95 (0.08) | 0.28 (0.11) | - | 0.81 (0.14) | 0.82 (0.12) | 0.80 (0.24) |
| | | MCIS | 0.76 (0.14) | 0.86 (0.11) | 0.91 (0.10) | 0.46 (0.13) | 0.97 (0.07) | 0.88 (0.10) | - | 0.85 (0.18) |

Table S7: Correlation of PVS volume, for each algorithm on each dataset. LO: Leave Out, 5FCV: Five fold cross validation. LOSOCV: Leave Out One Site Cross Validation. Average (standard deviation) is mean across datasets for which correlation was calculated.

| Region | Algorithm | Training Dataset | Internal Validation (5FCV) | ADNI (n=10) | AF (n=10) | MCIS (n=7) | Average |
|---|---|---|---|---|---|---|---|
| White Matter | SHIVA | - | - | 0.51 | 0.95 | 0.95 | 0.80 (0.25) |
| | WPSS | - | - | 0.84 | 0.84 | 0.91 | 0.86 (0.04) |
| | PINGU | All Sites | 0.69 | - | - | - | - |
| | PINGU | LO ADNI | 0.66 | 0.73 | - | - | 0.82 (0.13) |
| | | LO AF | 0.74 | - | 0.76 | - | |
| | | LO ASC | 0.63 | - | - | - | |
| | | LO HBA | 0.71 | - | - | - | |
| | | LO FTD | 0.68 | - | - | - | |
| | | LO MCIS | 0.68 | - | - | 0.98 | |
| | PINGU | ADNI | 0.74 | - | 0.77 | 0.90 | 0.84 (0.09) |
| | | AF | 0.78 | 0.68 | - | 0.97 | 0.83 (0.21) |
| | | HBA | - | 0.60 | 0.81 | 0.83 | 0.75 (0.13) |
| | | MCIS | 0.98 | 0.73 | 0.83 | - | 0.78 (0.07) |
| Basal Ganglia | SHIVA | - | - | 0.63 | 0.41 | -0.70 | 0.11 (0.71) |
| | WPSS | - | - | 0.49 | 0.82 | 0.37 | 0.56 (0.23) |
| | PINGU | All Sites | 0.82 | - | - | - | - |
| | PINGU | LO ADNI | 0.85 | 0.76 | - | - | 0.78 (0.12) |
| | | LO AF | 0.87 | - | 0.67 | - | |
| | | LO ASC | 0.67 | - | - | - | |
| | | LO HBA | 0.73 | - | - | - | |
| | | LO FTD | 0.79 | - | - | - | |
| | | LO MCIS | 0.75 | - | - | 0.91 | |
| | PINGU | ADNI | 0.87 | - | 0.72 | 0.86 | 0.79 (0.10) |
| | | AF | 0.53 | -0.53 | - | 0.88 | 0.18 (1.00) |
| | | HBA | - | 0.77 | 0.34 | 0.46 | 0.52 (0.22) |
| | | MCIS | 0.85 | 0.59 | 0.60 | - | 0.60 (0.01) |

Table S8: Correlation of PVS numbers, for each algorithm on each dataset. LO: Leave Out, 5FCV: Five fold cross validation. LOSOCV: Leave Out One Site Cross Validation. Average (standard deviation) is mean across datasets for which correlation was calculated.

| Region | Algorithm | Training Dataset | Internal Validation (5FCV) | ADNI | AF | MCIS | Average |
|---|---|---|---|---|---|---|---|
| White Matter | SHIVA | - | - | 0.48 | 0.87 | 0.91 | 0.75 (0.24) |
| | WPSS | - | - | 0.69 | 0.58 | 0.20 | 0.49 (0.26) |
| | PINGU | All Sites | 0.76 | - | - | - | - |
| | PINGU | LO ADNI | 0.71 | 0.71 | - | - | 0.84 (0.15) |
| | | LO AF | 0.79 | - | 0.81 | - | |
| | | LO ASC | 0.79 | - | - | - | |
| | | LO HBA | 0.84 | - | - | - | |
| | | LO FTD | 0.80 | - | - | - | |
| | | LO MCIS | 0.33 | - | - | 1.00 | |
| | PINGU | ADNI | 0.80 | - | 0.70 | 0.90 | 0.80 (0.14) |
| | | AF | 0.61 | 0.65 | - | 0.99 | 0.82 (0.24) |
| | | HBA | -0.17 | 0.78 | 0.82 | 0.82 | 0.81 (0.02) |
| | | MCIS | 0.99 | 0.67 | 0.82 | - | 0.75 (0.11) |
| Basal Ganglia | SHIVA | - | - | 0.03 | 0.39 | -0.41 | 0.00 (0.40) |
| | WPSS | - | - | 0.49 | 0.33 | 0.20 | 0.34 (0.15) |
| | PINGU | All Sites | 0.68 | - | - | - | - |
| | PINGU | LO ADNI | 0.65 | 0.53 | - | - | 0.46 (0.09) |
| | | LO AF | 0.75 | - | 0.36 | - | |
| | | LO ASC | 0.73 | - | - | - | |
| | | LO HBA | 0.83 | - | - | - | |
| | | LO FTD | 0.74 | - | - | - | |
| | | LO MCIS | 0.06 | - | - | 0.48 | |
| | PINGU | ADNI | 0.80 | - | 0.58 | 0.98 | 0.78 (0.28) |
| | | AF | 0.22 | 0.67 | - | 0.47 | 0.57 (0.14) |
| | | HBA | 0.06 | 0.59 | 0.43 | 0.84 | 0.62 (0.21) |
| | | MCIS | 0.89 | 0.48 | 0.43 | - | 0.45 (0.04) |

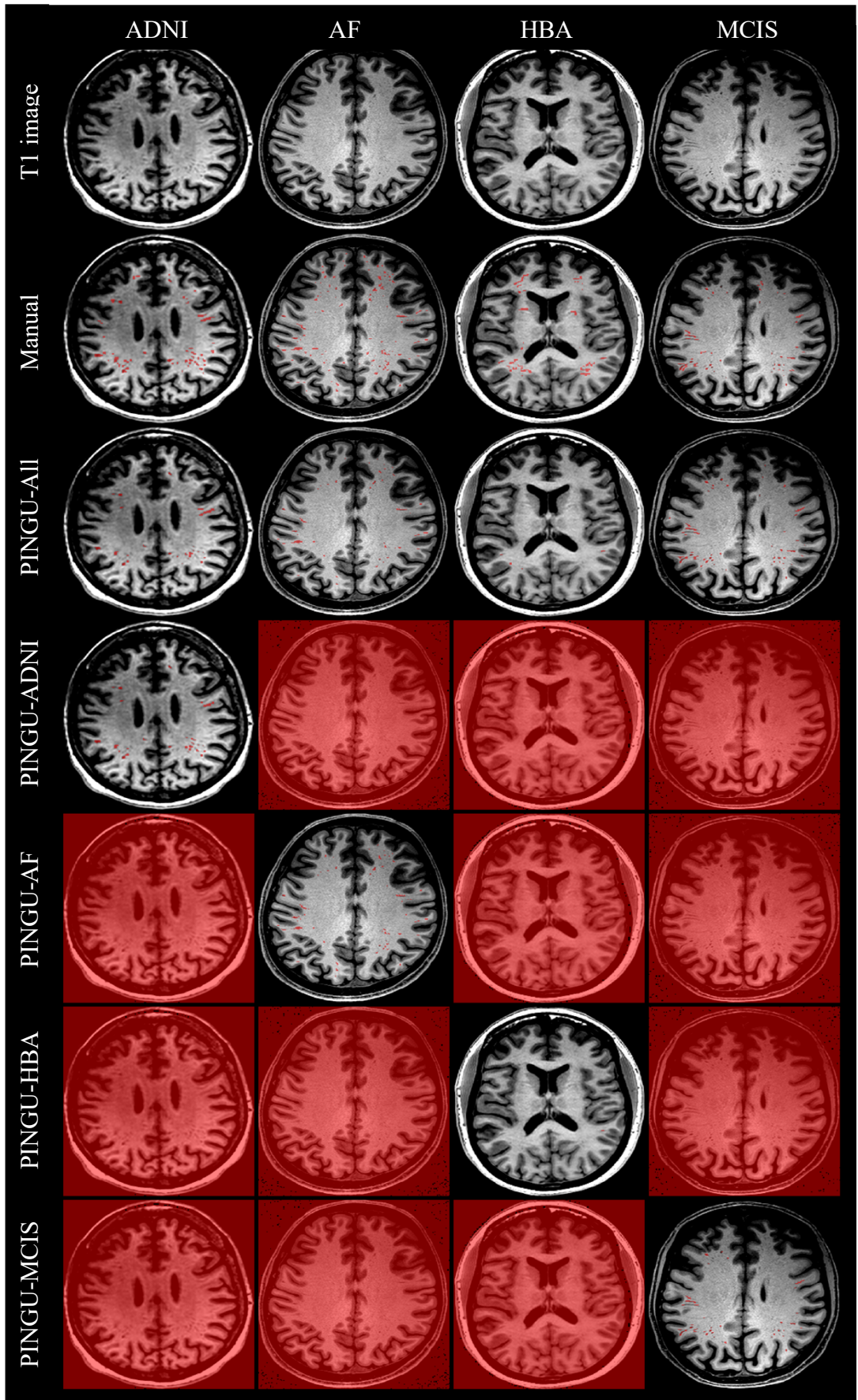

Figure S1: Internal validation: Overlays of segmentations of PINGU-All, and PINGU trained on each dataset individually in the white matter.

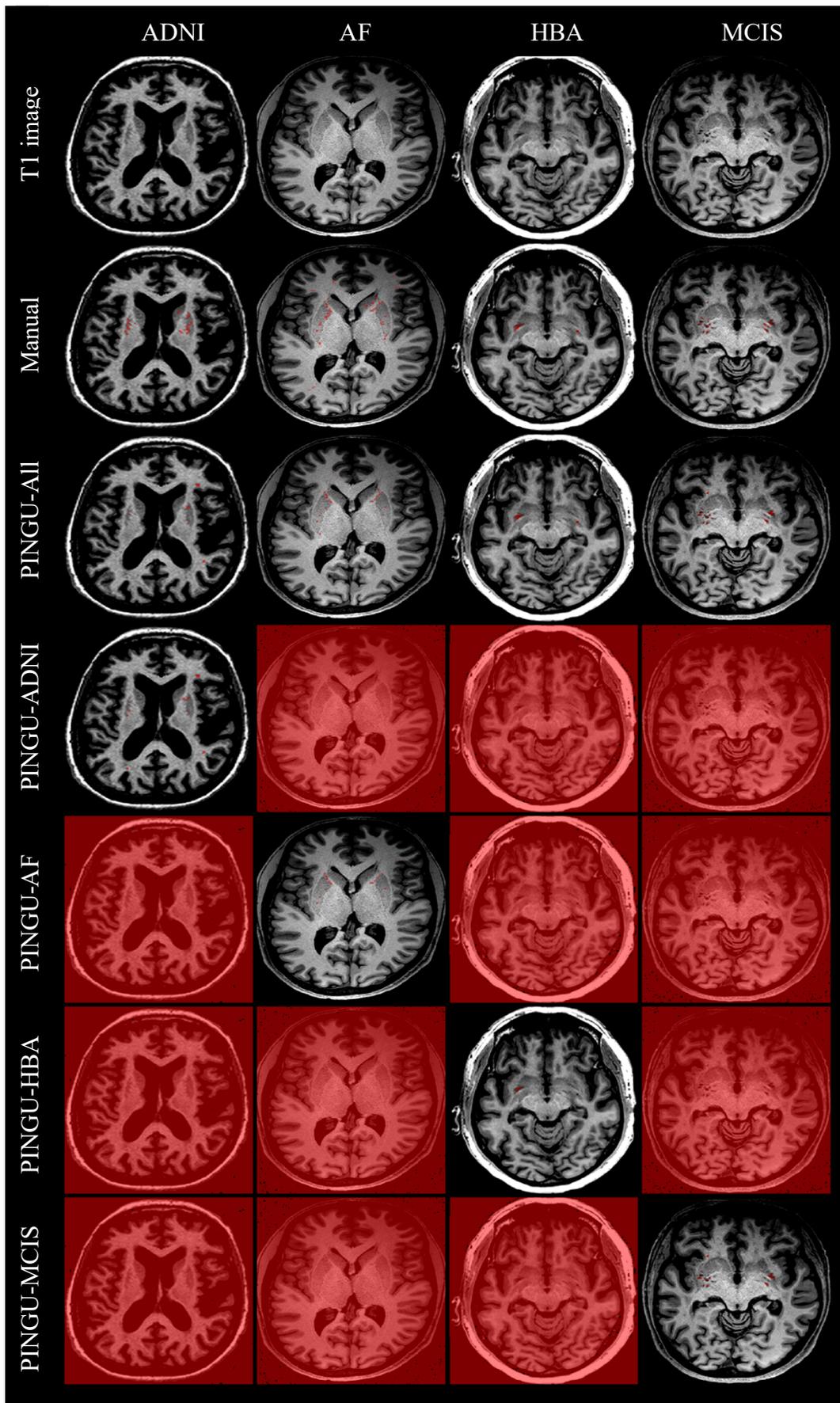

Figure S2: Internal validation: Overlays of segmentations of PINGU-All, and PINGU trained on each dataset individually in the Basal Ganglia.

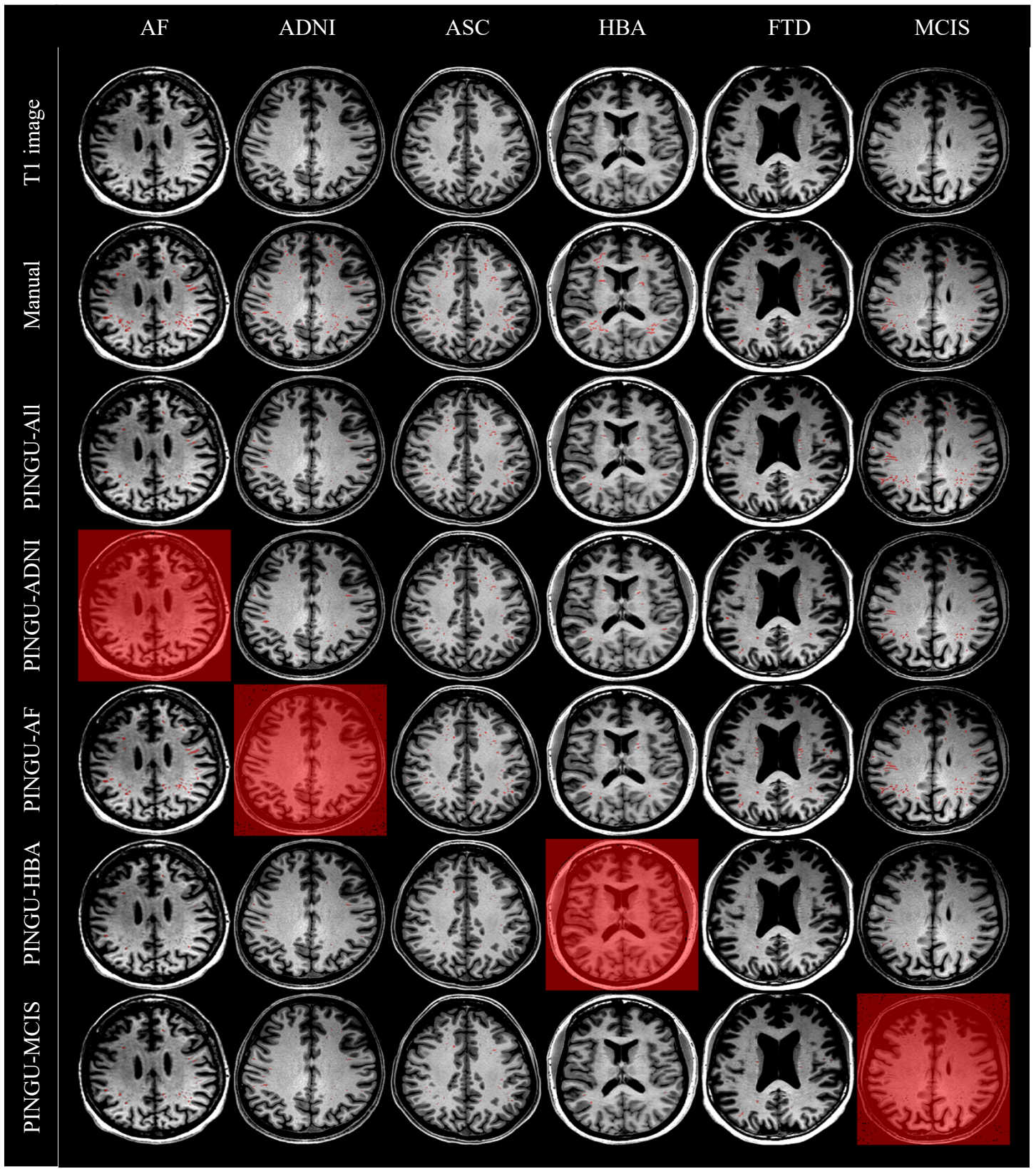

Figure S3: External validation: Overlays of segmentations of PINGU-LOSO, and PINGU trained on each dataset individually in the white matter.

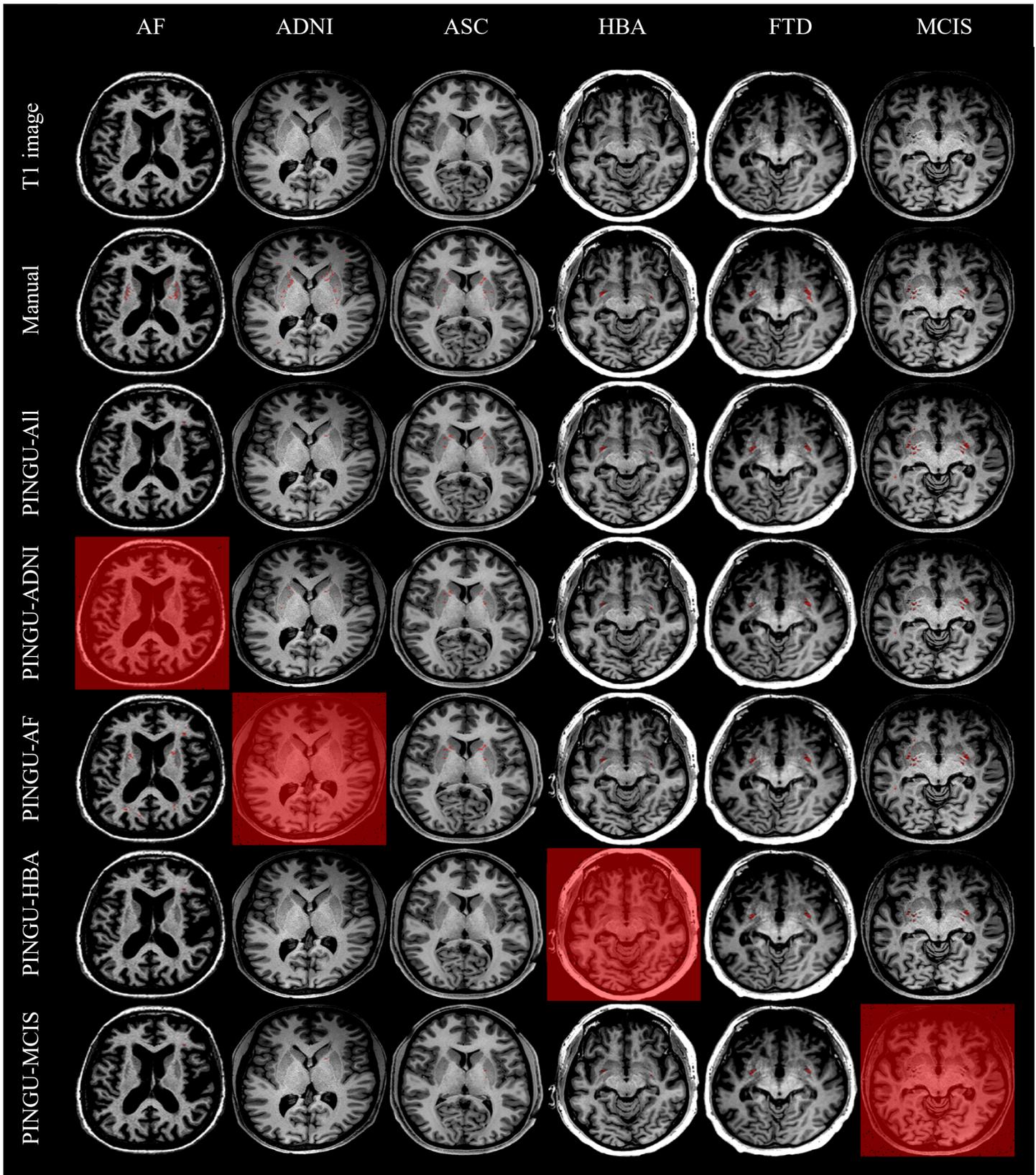

Figure S4: External validation: Overlays of segmentations of PINGU-LOSO, and PINGU trained on each dataset individually in the Basal Ganglia